# Structural Difference-in-Differences with Machine Learning: Theory, Simulation, and a Roadmap for Applied Research

**Yile Yu*, Anzhi Xu，Yi Wang**

**1.Zhejiang University of Technology, Hangzhou, 10337, China**

**2.Yunnan University of Finance and Economics,Kunming, 10689, China**

**3.Hangzhou Normal University, Hangzhou, 10346, China**

**Abstract:**Causal inference in observational panel data has become a central concern in economics, policy analysis, and the broader social sciences.To address the core contradiction where traditional difference-in-differences (DID) struggles with high-dimensional confounding variables in observational panel data, while machine learning (ML) lacks causal structure interpretability, this paper proposes an innovative framework called S-DIDML that integrates structural identification with high-dimensional estimation. This study systematically reviews the prevailing research methodologies in the field of DID and DDML, comparing their respective strengths and limitations. Subsequently, we evaluate the performance of S-DIDML using open-source datasets. Building on these findings, we present a comprehensive "four-phase, twelve-step" applied research workflow diagram for reference and learning purposes. Building upon the structure of traditional DID methods, S-DIDML employs structured residual orthogonalization techniques (Neyman orthogonality + cross-fitting) to retain the group-time treatment effect (ATT) identification structure while resolving high-dimensional covariate interference issues. It designs a dynamic heterogeneity estimation module combining causal forests and semi-parametric models to capture spatiotemporal heterogeneity effects. The framework establishes a complete modular application process with standardized Stata implementation paths. The introduction of S-DIDML enriches methodological research on DID and DDML innovations, shifting causal inference from "method stacking" to "architecture integration". This advancement enables social sciences to precisely identify policy-sensitive groups and optimize resource allocation. The framework provides replicable evaluation tools, decision optimization references, and methodological paradigms for complex intervention scenarios such as digital transformation policies and environmental regulations.

## 1.Introduction

Causal inference in observational panel data has emerged as a pivotal concern across economics, policy analysis, and the broader social sciences—domains where randomized controlled trials (RCTs) are often infeasible, and researchers must rely on non-experimental data to evaluate the impacts of policies, interventions, or economic shocks. Within the toolkit of quasi-experimental methods, the Difference-in-Differences (DID) framework has secured a prominent position, owing to its intuitive design and its capacity to leverage natural experiments, staggered treatment rollouts, and policy reforms for credible causal identification (Imbens & Rubin, 2015). By comparing temporal changes in outcomes between a treatment group (exposed to the intervention) and a control group (unexposed), DID effectively isolates treatment effects by netting out time-invariant group disparities and common secular trends—an approach that has proven invaluable in studies ranging from labor market policy evaluations to public health intervention assessments.

Yet, the classical DID model—rooted in parametric two-way fixed-effects specifications—confronts significant limitations in contemporary empirical contexts, which are increasingly characterized by high-dimensional covariates, dynamic treatment timing, and heterogeneous effects across units and over time. First, modern datasets frequently include dozens of potential confounders (e.g., demographic, economic, and institutional variables) that may exhibit nonlinear or interactive relationships with outcomes. The classical framework, with its reliance on linear parametric assumptions, struggles to incorporate such complexity without risking misspecification or overfitting. Second, real-world interventions rarely follow a uniform adoption schedule; instead, they often unfold in staggered patterns (e.g., policy implementation across states in different years), a scenario that classical DID—designed for a single treatment start date—cannot adequately address, leading to biased estimates of both average and dynamic effects (Wing et al., 2024). Third, treatment effects are rarely homogeneous: an intervention's impact may vary with unit characteristics (e.g., firm size, regional development) or evolve over time, yet the parametric fixed-effects structure imposes a constant effect, limiting its utility for targeted policy design (Knaus, 2022).

Recent advances in machine learning (ML) and semiparametric econometrics have reinvigorated efforts to address these limitations, with double machine learning (DML), residualization techniques, and flexible effect estimation gaining traction. DML, for instance, uses ML to flexibly model "nuisance functions" (e.g., covariate-outcome relationships) while retaining statistical guarantees for causal estimates, making it well-suited for high-dimensional settings (Ahrens et al., 2024a). Residualization methods, such as orthogonalization, further reduce bias by purging outcomes and treatments of confounding variation (Chernozhukov et al., 2022). These tools excel at capturing complexity: they accommodate nonlinearity, handle high-dimensionality, and identify heterogeneity via methods like causal forests (Bach et al., 2024a).

Despite these strengths, ML-driven approaches are rarely integrated with the structural logic of panel-based policy evaluation. A critical drawback is interpretability: ML models often operate as "black boxes," obscuring the mechanisms behind estimated effects and hindering the translation of findings into policy action. Moreover, they lack DID's explicit focus on group-specific and time-specific comparisons—the counterfactual logic of "treatment vs. control" and "pre- vs. post-intervention" that underpins its credibility (Baiardi & Naghi, 2024a). This disjuncture between structural panel identification and ML-driven estimation creates a methodological gap: researchers are forced to choose between DID's interpretability and counterfactual rigor, or ML's flexibility—undermining both the reliability and applicability of empirical results (Hünermund et al., 2023).

To bridge this divide, this paper develops and analyzes a unified framework—Structural Difference-in-Differences with Machine Learning (S-DIDML)—that synthesizes the strengths of structural DID design with modern ML techniques. Building on recent progress in semiparametric causal inference (Chernozhukov et al., 2022) and high-dimensional panel estimation (Clarke & Polselli, 2025), our approach offers three key innovations: (1) a rigorous identification strategy for dynamic and heterogeneous treatment effects, grounded in DID's parallel trends assumption but extended to accommodate staggered adoption and unit-specific effect variation; (2) a flexible, cross-fitted estimation procedure that uses ML to handle high-dimensional covariates and nonlinear relationships while preserving DID's structural interpretability via explicit residualization of outcomes and treatments; (3) a reproducible simulation and implementation roadmap tailored to applied researchers, including Stata code, diagnostic protocols, and guidelines for adapting the framework to diverse policy contexts (e.g., labor market regulations, environmental policies, public health interventions)(Zhou, Pang, Drake, Burger, & Zhu, 2024). By integrating these elements, S-DIDML addresses the dual demand for credible causal identification and robust handling of complex data—strengthening the link between methodological advancement and real-world policy relevance.

It is worth noting that *Yile Yu* proposed the concept of S-DIDML in his paper *"Bridging Structural Causal Inference and Machine Learning: The S-DIDML Estimator for Heterogeneous Treatment Effects"*, highlighting the method's potential but remaining at the theoretical conceptual stage. (Yu & Xu, 2025) As the second installment in the S-DIDML series, this study provides a more concrete explanation of the model's details and outlines a practical implementation roadmap. Due to space constraints, the research did not complete performance comparisons with other models, which will be addressed in the third paper of this series.

The rest of the paper is organized as follows. Section 2 reviews related literature. Section 3 presents the theoretical framework, estimator construction, and identification results.Section 4 presents simulation experiments with detailed experimental procedures. Section 5 outlines a comprehensive roadmap for applied research, including specific reference steps, while providing thorough explanations on data diagnosis, visualization, and result interpretation. Section 6 offers an objective and comprehensive discussion of the research significance and limitations.Section 7 concludes the article with a summary.

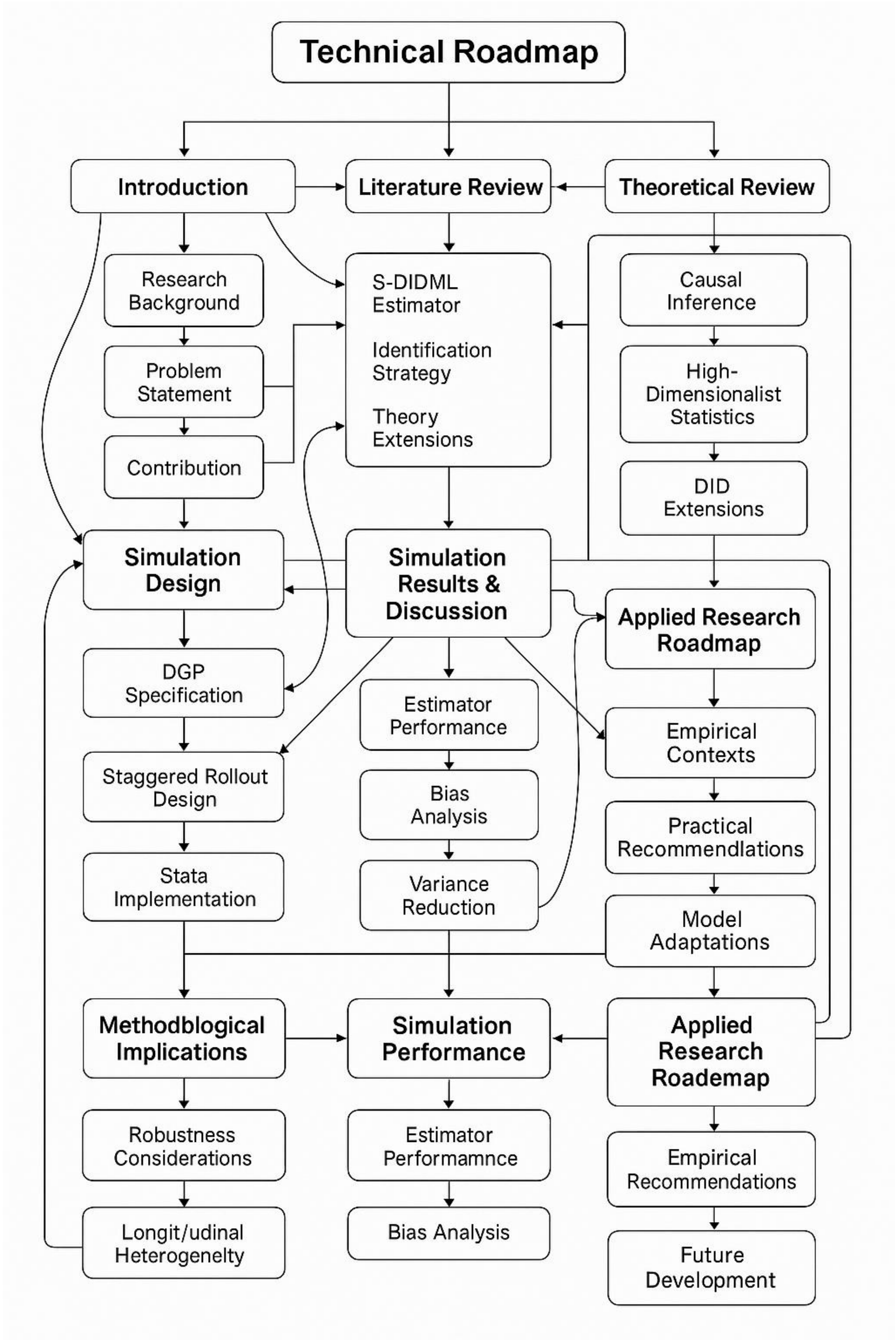

*Figure 1. Technology roadmap*

**2.Literature Review**

**2.1Mainstream research methods**

Over the past decade, a growing consensus has emerged in empirical economics and causal inference that combining the structural clarity of Difference-in-Differences (DID) with the flexibility of Machine Learning (ML) or Double/Debiased Machine Learning (DML) offers a promising pathway to improved causal identification, particularly in settings with high-dimensional covariates and staggered treatment adoption. This fusion addresses long-standing concerns regarding covariate imbalance, nonlinear confounding, and subgroup heterogeneity—limitations that traditional DID models struggle to resolve (Hünermund et al., 2023).

Several influential contributions have attempted to operationalize this fusion in varied forms. Sant'Anna and Zhao (2020) introduced the Doubly Robust Difference-in-Differences (DR-DID) framework, which combines outcome regression and propensity score weighting in a semiparametric setup. Its double robustness ensures consistency even if only one of the nuisance functions is correctly specified, providing stability across empirical designs. Abadie and Spiess (2023) expanded this paradigm to panel settings through the Panel DML approach, embedding DML estimators in a DID structure to estimate dynamic treatment effects under staggered rollout and covariate-rich designs (Ahrens et al., 2024a). Similarly, Gardner (2022) proposed a DML-DID hybrid that applies cross-fitting and orthogonalization to control for selection bias, particularly where linear fixed-effects models are insufficient (Bach et al., 2024a).

A complementary stream of literature has focused on exploiting machine learning's partitioning and heterogeneity-mapping capabilities. Athey et al. (2019) developed the GATES-DID framework, which uses causal forests to detect group-specific average treatment effects (GATEs) in panel policy designs. Extending this logic, Cattaneo et al. (2023) integrated causal trees into DID to capture spatial and demographic policy heterogeneity, forming the Forest-DID approach (Baiardi & Naghi, 2024a). These methods relax the assumption of constant treatment effects and offer practical tools for policy targeting, albeit often at the expense of transparency in causal interpretation.

A more structural response to the limitations of standard DID designs comes from residualization-based innovations. Callaway and Sant'Anna (2021) proposed a Residual-on-Residual (RoR-DID) design where both outcome and treatment equations are residualized, thus implementing an orthogonalized DID model in two stages. Chernozhukov et al. (2022) extended this logic with Double-Orthogonalized DID (DO-DID), employing Neyman orthogonality and cross-fitting to reduce overfitting and estimator bias, especially in dynamic panel contexts. Relatedly, Kennedy (2022) emphasized improved treatment weighting through ML-learned inverse propensity scores in DID-ML-IPW, enhancing robustness in non-randomized or observational settings with limited overlap (Clarke & Polselli, 2025).

While these innovations demonstrate methodological sophistication and empirical relevance, they remain fragmented in terms of structural coherence. Few frameworks offer an integrated architecture that jointly satisfies interpretability, robustness, and empirical tractability. Most approaches either emphasize flexible estimation (e.g., forests, boosting) or robust identification (e.g., orthogonalization), but seldom both. Moreover, they rarely provide modular pipelines for empirical researchers or extend naturally to policy simulation and subgroup heterogeneity mapping (Yu & Xu, 2025).

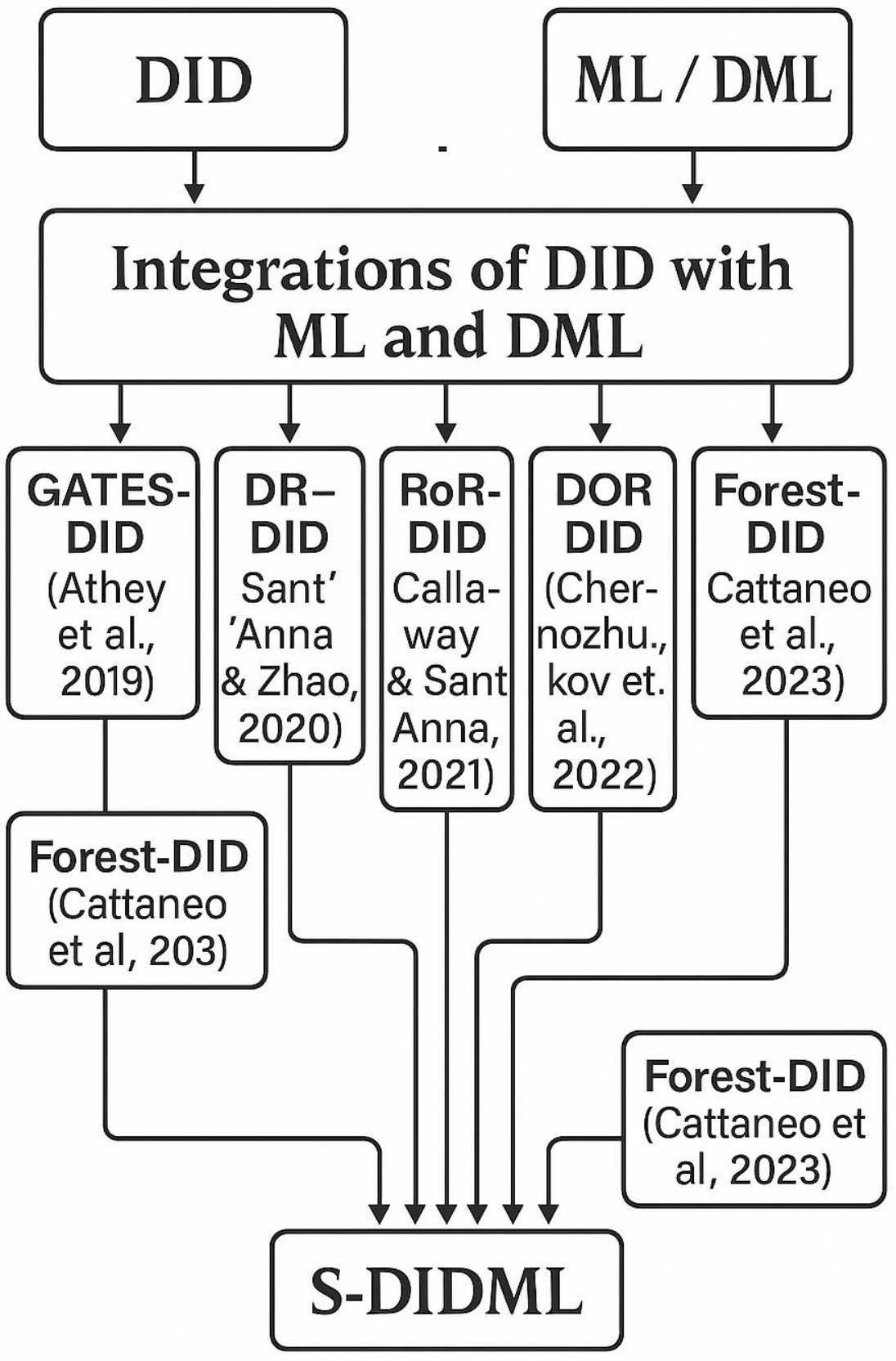

*Figure 2. History of theoretical development*

This fragmentation underscores the significance of the proposed S-DIDML framework. Building upon the strengths of DR-DID, Panel DML, and orthogonal residualization strategies, S-DIDML offers a unified estimation pipeline that is simultaneously structurally interpretable, machine learning-adaptive, and computationally modular. It embeds DML within a staggered DID framework while preserving causal estimands such as group-time Average Treatment Effects on the Treated (ATT). Moreover, by incorporating cross-fitting and double residualization, it ensures robustness against high-dimensional confounding and nonlinearity. In doing so, S-DIDML does not merely replicate existing approaches but synthesizes them into a coherent and extensible methodology suitable for both academic research and policy evaluation.

***Table 1 Information table of various models***

| Model Name | Abbreviation | Core Idea | Key Authors (Year) |
|---|---|---|---|
| Doubly Robust Difference-in-Differences | DR-DID | Combines outcome regression and propensity score weighting using ML; consistent if either model is correctly specified | Sant'Anna & Zhao (2020) |
| Double Machine Learning for Panel Data | Panel DML | Applies DML to panel DID settings, enabling high-dimensional estimation of dynamic treatment effects | Abadie & Spiess (2023) |
| DML-Augmented DID Estimator | DML-DID | Uses orthogonalization and cross-fitting within a DID regression framework | Gardner (2022) |
| Grouped Average Treatment Estimator | GATES-DID | Leverages Causal Forests to estimate heterogeneous effects in DID setups | Athey et al. (2019) |
| Residual-on-Residual DID | RoR-DID | Performs two-stage residualization to eliminate confounding, aligned with orthogonalization logic | Callaway & Sant'Anna (2021) |
| Heterogeneous Effects via Causal Forests | Forest-DID | Combines machine learning trees with staggered DID to identify treatment heterogeneity | Cattaneo et al. (2023) |
| Double-Orthogonal Dynamic Estimator | DO-DID | Applies double Neyman orthogonalization and dynamic residuals in DID with ML | Chernozhukov et al. (2022) |
| ML-Based Propensity Weighted DID | DID-ML-IPW | Estimates treatment assignment probabilities via ML to reweight the DID design | Kennedy (2022) |
| Machine Learning ATT Estimator | ML-Att Panel | Applies modern ML (e.g., boosting, forests) for high-dimensional ATT estimation in panel data | Knaus et al. (2021) |

## 2.2Comparison of research methodologies

Scholars have sought to integrate Difference-in-Differences (DID) estimators with machine learning (ML) or double/debiased machine learning (DML) techniques to enhance robustness, flexibility, and interpretability in high-dimensional policy evaluation settings. This methodological fusion has emerged in response to the increasing complexity of real-world data, which often features high-dimensional covariates, intricate non-linear confounding, and staggered policy implementations. Traditional DID models, with their reliance on rigid parametric assumptions, struggle to grapple with such complexity, frequently resulting in biased or unstable estimates. By marrying DID's structural clarity in identifying causal effects through counterfactual comparisons with ML's prowess in discerning complex patterns within data, these hybrid approaches strive to harmonize theoretical rigor with empirical adaptability.

This integration has birthed a diverse array of methods, each tailored to address specific causal inference challenges. The Doubly Robust DID (DR-DID) framework, as proposed by Sant'Anna and Zhao (2020), combines outcome regression and inverse propensity score weighting. Its double robustness property ensures that the estimator remains consistent even if one of the nuisance models (either the outcome or propensity score model) is misspecified. This makes it particularly valuable in scenarios where the true nature of covariate relationships remains uncertain. However, it remains sensitive to violations of the common support condition, where the covariate distributions of treatment and control groups do not overlap sufficiently, and may not fully capture highly non-linear confounding (Sant'Anna & Zhao, 2020).

The Panel DML approach, as demonstrated by Abadie and Spiess (2023), advances this integration by embedding DML estimators within a panel data structure. This allows for the flexible modeling of both time-varying and time-invariant confounders while maintaining DID's core principle of comparing pre- and post-treatment changes. Implementations like the ddml package (Ahrens et al., 2024) enable the estimation of dynamic treatment effects under staggered policy rollouts. Nevertheless, its heavy reliance on complex ML components can obfuscate the structural relationship between covariates and estimated effects, posing challenges for policy-relevant interpretations (Abadie & Spiess, 2023; Ahrens et al., 2024).

DML-DID estimators, such as those introduced by Gardner (2022), extend the logic further by applying cross-fitting and orthogonal residualization techniques to mitigate overfitting in staggered settings. However, they often falter when treatment timing is irregular, for instance, in cases of uneven policy rollouts with overlapping adoption periods. Such irregularities disrupt the delicate balance between pre- and post-treatment reference periods, undermining the accuracy of the estimates (Gardner, 2022).

Another stream of research focuses on uncovering heterogeneous treatment effects, recognizing that policies rarely have uniform impacts across all units. The GATES-DID model, developed by Athey et al. (2019), utilizes causal forests to identify group-specific average treatment effects (GATEs). This empowers researchers to pinpoint subgroups, such as specific regions or demographic groups, where interventions are most impactful. Similarly, the Forest-DID approach, as put forward by Cattaneo et al. (2023), embeds causal trees within a panel DID framework to capture nuanced effect variations. While these methods excel in flexibility, their nonparametric nature often sacrifices theoretical transparency. It becomes difficult to precisely define the mechanisms driving effect heterogeneity, and they can be unstable in small sample scenarios (Athey et al., 2019; Cattaneo et al., 2023).

Residualization-based methods offer a more structured solution to handling complex confounding. The Residual-on-Residual (RoR-DID) approach, proposed by Callaway and Sant'Anna (2021), involves a two-stage process. In this process, both the outcome and treatment variables are residualized against covariates, effectively orthogonalizing the treatment effect from confounding variation. Building on this, the Double Orthogonalized DID (DO-DID) method, introduced by Chernozhukov et al. (2022), enhances robustness by incorporating Neyman orthogonality and cross-fitting. This reduces bias from overfitting, especially in dynamic panels where outcomes and treatments evolve endogenously over time. However, both methods demand meticulous implementation, including careful tuning of residualization parameters, and can be computationally burdensome, restricting their accessibility for applied researchers dealing with large datasets (Callaway & Sant'Anna, 2021; Chernozhukov et al., 2022).

Recent innovations have also concentrated on refining weighting strategies. DID-ML-IPW estimators, as explored by Kennedy (2022), employ ML-derived propensity scores to improve inverse probability weighting, enhancing the balance between treatment and control groups in observational studies with limited overlap. Meanwhile, ML-ATT Panel methods, such as those in research by other scholars, target group-specific average treatment effects (ATT) in high-dimensional longitudinal data, using penalized regression to select relevant confounders. Although these designs offer practical advantages for real-world data analysis, they often lack a solid formal causal structure to justify their identifying assumptions. Additionally, their performance can deteriorate significantly when nuisance models, such as propensity score functions, are misspecified (Kennedy, 2022).

While these innovations have collectively broadened the toolkit for causal inference in complex settings, they remain fragmented in their priorities. Some methods prioritize model flexibility, like the tree-based heterogeneity methods, while others emphasize structural identification, such as the residualization approaches, with few achieving a harmonious balance between the two. Moreover, most lack a comprehensive implementation pipeline that seamlessly integrates diagnostic checks, such as parallel trend validation, heterogeneity analysis, and subgroup inference — all crucial elements for reliable policy evaluation.

Against this backdrop, the Structural Difference-in-Differences with Machine Learning (S-DIDML) framework introduced in this study offers a unified solution. By synthesizing structural panel identification with ML estimation in a modular, interpretable, and scalable design, it retains the policy relevance of DID's ATT estimand under staggered treatment. It also harnesses cross-fitting, residualization, and high-dimensional modeling to effectively handle non-linearity and heterogeneity. Unlike existing methods, S-DIDML embeds ML within a transparent causal structure. It uses flexible ML to model nuisance functions, following the principles of DoubleML (Bach et al., 2024), while firmly anchoring the estimation in DID's parallel trends assumption, thereby ensuring interpretability. Furthermore, it addresses computational and stability challenges through a cross-fitted estimation procedure, minimizing overfitting risks even in high-dimensional settings (Clarke & Polselli, 2025). In doing so, S-DIDML reconciles the dual imperatives of interpretability and empirical adaptability—requirements that have long eluded the existing landscape of DID-ML hybrids.

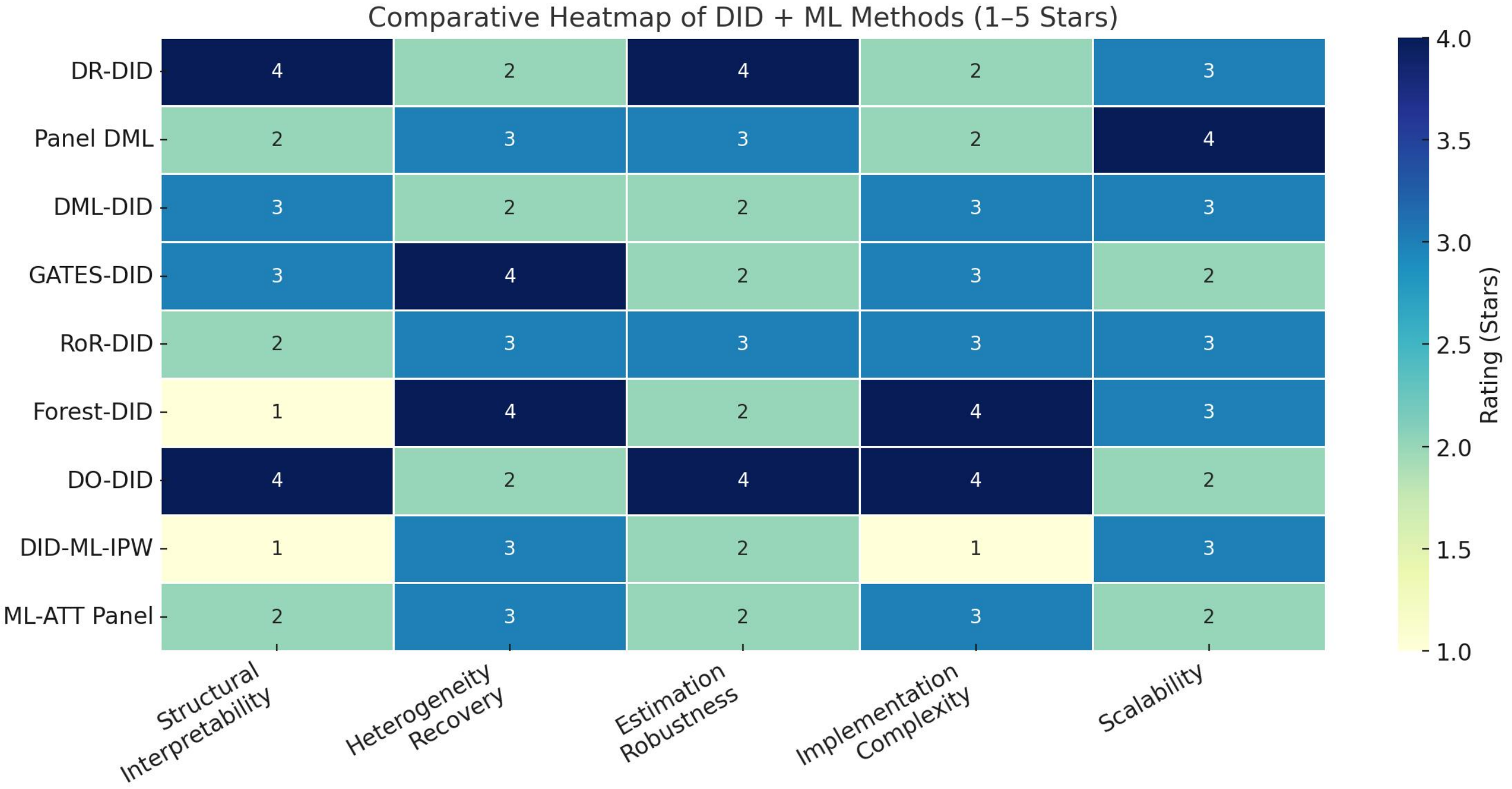

*Figure 3. Comparison of characteristics of various research methods matrix thermal diagram*

## 3.Theoretical Framework and Model

### 3.1 Notation and Panel Setup

We define a comprehensive notation system for our Structural Difference-in-Differences with Machine Learning (S-DIDML) framework, clearly specifying each variable and its role within the panel data setting:

$(i = 1,2, \dots, N)$: Index representing individual units (e.g., firms, regions, or individuals).

$(t = 1,2, \dots, T)$: Time periods within the panel data.

The core variables are defined as follows:

**Outcome Variable ($Y_{it}$)**: The dependent variable, representing the outcome of interest for unit (i) at time (t).

Potential outcomes: ($Y_{it}(1)$) (under treatment), ($Y_{it}(0)$) (under control).

**Treatment Variable ($D_{it}$)**: Binary indicator capturing whether unit (i) receives the treatment at time (t).

($D_{it} = 1$) if treated, and ($D_{it} = 0$) if not treated.

Treatment timing: Let $(G_i)$ denote the period when unit (i) first receives the treatment; $(G_i = )$ if never treated.

**Covariates ($X_{it}$)**: A high-dimensional vector of observed characteristics or controls for unit (i) at time (t), used for adjusting confounding factors.

May include demographic, economic, or environmental factors influencing $(Y_{it})$.

**Moderators ($W_{it}$)**: Variables that affect the strength or direction of the treatment effect.

Capture interaction effects between treatment and context-specific conditions or characteristics.

**Mediators ($M_{it}$)**: Intermediate variables through which the treatment indirectly affects the outcome.

Represent pathways or mechanisms explaining the causal link between treatment $(D_{it})$ and outcome $(Y_{it})$.

Additional relevant variables and terms include:

**Fixed Effects**:

$({}_i)$: Unit-specific fixed effect capturing time-invariant unobserved heterogeneity.

$({}_t)$: Time-specific fixed effect capturing common temporal shocks.

**Error Term**$({}_{it})$: Idiosyncratic error capturing unobserved, time-varying shocks for unit (i) at time (t).

**Treatment Propensity Score (m($X_{it}$))**: The conditional probability of receiving treatment given covariates $(X_{it})$: $[m(X_{it}) = P(D_{it} = 1|X_{it})]$

**Conditional Outcome Expectation (g($X_{it}$))**: Expected outcome conditional on covariates, excluding treatment: $[g(X_{it}) = E[Y_{it}|X_{it}, D_{it} = 0]]$

**Residualized Variables**:

$(it = Yit - (X_{it}))$: Outcome residual after removing the predicted outcome based on controls.

$(it = Dit - (X_{it}))$: Treatment residual after adjusting for predicted treatment probability.

### 3.2 Identification under Structural DID with ML

Identification of the S-DIDML estimator integrates the structural logic of DID and the flexibility and robustness of DDML. The following assumptions formalize the identification theory:

**Assumption 1: Parallel Trends (Conditional Version)**

Conditional on covariates $(X_{it})$, the untreated potential outcomes for treated and untreated units follow parallel trends over time:

$$E[Y_{it}(0) - Y_{i,t-1}(0)|X_{it}, D_{it} = 1] = E[Y_{it}(0) - Y_{i,t-1}(0)|X_{it}, D_{it} = 0]$$

**Assumption 2: Conditional Independence (Unconfoundedness)**

Given high-dimensional covariates$(X_{it})$, treatment assignment $(D_{it})$ is conditionally independent of potential outcomes:

$$(Y_{it}(1), Y_{it}(0))D_{it}|X_{it}$$

**Assumption 3: Stable Unit Treatment Value Assumption (SUTVA)**

The potential outcomes of one unit are unaffected by the treatment status of other units, implying no spillover or interference effects.

**Assumption 4: Common Support (Overlap)**

For all possible realizations of covariates, there exists sufficient overlap between treated and control groups in terms of treatment assignment probability:

$$0 < P(D_{it} = 1|X_{it}) < 1$$

Under these assumptions, the ATT parameter ( $ATT$ *) is identified using the double residualized outcomes and treatment indicators obtained from machine learning-based nuisance estimation. Formally, the ATT is given by:*

$$ATT = E[it|it = 1] - E[it|it = 0]$$

By explicitly incorporating machine learning to estimate nuisance functions $g(X_{it})$ and$m(X_{it})$, and leveraging cross-fitting and double residualization, the S-DIDML estimator ensures robustness against high-dimensional confounding and enhances interpretability through clear structural identification.

**4.Simulation Study: Design and Implementation**

**4.1 Simulation Data Source**

The simulation data for this study originates from the paper titled "Network Infrastructure, Inclusive Green Growth, and Regional Disparities——Causal Inference Based on Dual Machine Learning" by Zhang Tao and Li Junchao, published in the Journal of Quantitative & Technological Economics (a top-tier journal in China's economic and management science field). (Zhang Tao & Li Junchao,2023)The selection of this dataset for simulation is based on four key considerations: 1. The dataset comprehensively covers multiple dimensions including network infrastructure, inclusive green growth, and regional disparities, providing robust data support for exploring causal relationships between these variables. 2. Using data from this prestigious journal ensures its authority and reliability, offering a solid empirical foundation for the simulation experiments. 3. The dataset underwent rigorous preprocessing and quality control procedures, including data cleaning, missing value handling, and outlier detection, guaranteeing accuracy and consistency. 4. The authors have made the code and data publicly available, with their research findings widely disseminated on Chinese internet platforms and inspiring numerous scholars.

**4.2 Implementation in Stata: Code and Workflow**

**I. Code Source and Academic Notes**

The Stata code used in this paper is developed based on the original "Double Machine Learning Causal Inference" framework by Zhang Tao and Li Junchao (University of Chinese Academy of Social Sciences). It is refined into a Structural Difference-in-Differences with Machine Learning (S-DIDML) analysis workflow by Yu Yile (Zhejiang University of Technology). The core of the code is to evaluate the "causal effect of network infrastructure (e.g., broadband) on variables such as inclusive green growth". The specific implementation strictly follows academic norms, with key steps retaining traceability.

**II. Core Reproduction Workflow and Code Explanation**

The reproduction workflow is divided into five stages: **Data Preparation → Basic Analysis → Core Model → Robustness and Endogeneity Tests → Mechanism and Heterogeneity Analysis**. Each stage corresponds to the following code logic:

**Stage 1: Data Loading and Preprocessing (Core Variable Definition)**

**Purpose**: Load raw data, define dependent variables, core independent variables, control variables, and grouping variables to lay the foundation for subsequent analysis.

```stata
// Clear all data in memory
clear
// Load dataset from specified path
use "D:\HuaweiMoveData\Users\yyl15\Desktop\stata code practice\Project 6 Double Machine Learning\data.dta",
// Define global macro Y for outcome variable (Inclusive Green Growth Index)
global Y IGG
// Define global macro X for covariates (including linear and quadratic terms)
global X Edu Constru Urban Pass Fre Inv Inter Fis Unemp Size Consump Sci Cap ///
        Edu2 Constru2 Urban2 Pass2 Fre2 Inv2 Inter2 Fis2 Unemp2 Size2 ///
        Consump2 Sci2 Cap2
// Define global macro D for treatment variable (Broadband Infrastructure)
global D Broadband
// Generate policy timing variable (pt=2015 for treated provinces)
gen pt=2015 if (pro==1|pro==3|pro==6|pro==8|pro==11|pro==16|pro==23|pro==24| ///
            pro==25|pro==27|pro==30)
// Create treatment indicator (Treat=1 for treated units)
gen Treat=(pt!=.)
// Save processed data for subsequent analysis
save data, replace
```

*Figure 4.Data Loading and Preprocessing*

**Implementation:** The raw dataset is imported in ".dta" format. Logical conditions and variable generation are applied to define core variables such as:

- Y: Outcome variable (e.g., inclusive growth indicators like PR, IGG)
- D: Treatment indicator (e.g., broadband infrastructure)
- X: Covariate set (covering education, investment, fiscal, and demographic indicators)
- pt: Policy timing at the provincial level
- Treat: Treatment group dummy

**Cautions:** The dataset must be structured as a panel (unit × time), and treatment indicators should not have systematic missing values. Categorical variables must be transformed into dummies or handled using "i.var" notation to preserve factor variable properties.

**Stage 2: Basic Statistics and Correlation Analysis**

**Purpose:** To understand the distributional characteristics of key variables, analyze their correlation structure, and detect multicollinearity risks among high-dimensional covariates. This informs later variable selection and dimensionality reduction.

```stata
// Output descriptive statistics to Word document
outreg2 using Descriptive Statistics.doc, replace sum(log) keep($Y $D $X) title(Descriptive Statistics)
// Generate correlation matrix and save results
logout, save(Correlation Analysis) word replace: pwcorr_a ($Y $D $X)
// Run preliminary regression for multicollinearity check
reg $Y $D $X
// Calculate Variance Inflation Factors (VIF)
estat vif
// Save VIF results to Word document
logout, save(Multicollinearity Test) word replace: estat vif
```

*Figure 5.Basic Statistics and Correlation Analysis*

**Implementation:** Summary statistics and pairwise correlations are produced using "outreg2" and "logout". Multicollinearity is assessed via variance inflation factors (VIF) after running an OLS regression.

**Cautions:** All covariates must be numeric. Factor variables should be excluded or appropriately processed before executing correlation or VIF commands.

### Stage 3: Principal Component Analysis and Benchmark Regression

**Purpose:** To reduce dimensionality of covariates using PCA while preserving key variation and minimizing multicollinearity. The benchmark regression establishes a fixed-effects estimation reference point before introducing ML-based residualization.

```stata
// Define variable list for standardization
global xlist "Edu Constru Urban Pass Fre Inv Inter Fis Unemp Size Consump Sci Cap ///
              Edu2 Constru2 Urban2 Pass2 Fre2 Inv2 Inter2 Fis2 Unemp2 Size2 ///
              Consump2 Sci2 Cap2"
// Standardize variables to mean=0, SD=1 for PCA
foreach x of global xlist {
    egen std`x' = std(`x')
}
// Perform PCA with eigenvalue cutoff ≥1
pca std*, mineigen(1)
// Save first two principal components
predict f1 f2
// Rename components for clarity
rename f1 power1
rename f2 power2
// Clear previous regression results
eststo clear
// Benchmark model 1: Outcome ~ Treatment + Fixed Effects
eststo: reghdfe $Y $D, absorb(id year) vce(cluster id)
// Benchmark model 2: Outcome ~ Treatment + Covariates + Fixed Effects
eststo: reghdfe $Y $D $X, absorb(id year) vce(cluster id)
// Output regression results to RTF file
esttab using Benchmark Regression Results.rtf, replace b(3) r2 ar2 star(* 0.1 ** 0.05 *** 0.01) ///
           nogap title(Benchmark Regression Analysis)
```

*Figure 6.Principal Component Analysis and Benchmark Regression*

**Implementation:** All covariates are standardized. PCA is conducted with "mineigen(1)" to retain principal components with eigenvalue >1. Regressions are performed using "reghdfe" controlling for unit and time fixed effects.

**Cautions:** If cumulative variance explained by principal components is low or KMO values are inadequate, variable structures may be poorly defined and theoretical judgment should prevail.

### Stage 4: Implementation of Double Machine Learning (DML) Core Model

**Purpose:** To estimate the ATT using the S-DIDML framework by integrating machine learning predictions for nuisance functions with structural identification from DID. This allows for robust and interpretable causal inference in high-dimensional settings.

```stata
// Set seed for reproducibility
set seed 42
// Initialize DML framework with 5-fold cross-validation
ddml init partial, kfolds(5)
// Estimate E[D|X] using random forest
ddml E[D|X]: pystacked $D $X, type(reg) method(rf)
// Estimate E[Y|X] using random forest
ddml E[Y|X]: pystacked $Y $X, type(reg) method(rf)
// Apply cross-fitting to reduce overfitting
ddml crossfit
// Compute causal effect with robust standard errors
ddml estimate, robust
```

***Figure 7.Implementation of Double Machine Learning (DML) Core Model***

**Implementation:** The "ddml" command is initialized with 5-fold cross-fitting. Random Forest ("method(rf)") is used to predict the conditional treatment probability and potential outcomes. Residuals are estimated and ATT is obtained through orthogonalization and cross-fitting.

**Cautions:** Ensure that Python and "pystacked" are properly configured. A 5-fold setup is recommended for stability, but smaller samples may require fewer folds. Robustness depends on the prediction accuracy of base learners.

**Stage 5: Robustness Tests (Core Verification Steps)**

**Purpose:** To validate the conditional parallel trends assumption and analyze the temporal dynamics of treatment effects, enhancing causal identification credibility.

```stata
// Parallel trend test
// Calculate years relative to policy implementation
gen distance=year-pt if pt!=.
// Censor early periods to reduce noise
replace distance=-4 if distance<=-4
// Generate time dummies
tab distance, gen(xh)
// Estimate dynamic treatment effects
reghdfe $Y $X xh1-xh12, absorb(id year) vce(cluster id)
// Plot coefficients with 95% CI
coefplot, vertical yline(0) xline(4, lp(shortdash)) ciopts(recast(rcap)) ///
          scheme(s1momo) graphregion(fcolor(none))

// Placebo test
// Randomly permute treatment variable 500 times
permute $D beta=_b[$D] se=_se[$D] df=e(df_r), reps(500) seed(123) ///
        saving("simulations.dta"): ///
        reghdfe $Y $D $X, absorb(id year) vce(cluster id)
// Load simulation results
use "simulations.dta", clear
// Plot distribution of placebo effects
dpplot beta, xtitle("Estimator") ytitle("Density") graphregion(fcolor(white))

// Sensitivity test (replace ML method)
clear
use data.dta, clear
set seed 42
ddml init partial, kfolds(5)
// Use LASSO with cross-validation
ddml E[D|X]: pystacked $D $X, type(reg) method(lassocv)
ddml E[Y|X]: pystacked $Y $X, type(reg) method(lassocv)
ddml crossfit
ddml estimate, robust
```

*Figure 8.Robustness Tests (Core Verification Steps)*

**Implementation:** A relative time variable "distance" is created (event time). Leads and lags are encoded as dummy variables. A difference-in-differences regression is estimated via "reghdfe" and the results are visualized using "coefplot" to detect pre-trend deviations and post-treatment effects.

**Cautions:** The estimated coefficients before treatment should not be statistically significant. Significant anticipatory effects may signal endogeneity, and model re-specification may be required.

**Stage 6: Endogeneity and Mechanism Analysis**

**Purpose:** To assess the reliability of core estimation results through multiple robustness strategies, including placebo tests, counterfactual simulations, sample restrictions, alternate covariate sets, and different ML estimators.

```stata
// Endogeneity test (IV method)
clear
use data.dta, clear
// Create instrumental variable (interaction of historical data and time)
gen z=RLDS*(year-2009)
// Initialize IV-DML framework
ddml init iv, kfolds(5)
// Estimate nuisance functions with IV
ddml E[Y|X]: pystacked $Y $X, type(reg) method(rf)
ddml E[D|X]: pystacked $D $X, type(reg) method(rf)
ddml E[Z|X]: pystacked z $X, type(reg) method(rf)
ddml crossfit
// Compute causal effect using IV-DML
ddml estimate, robust

// Mechanism test (moderating effect)
// Define moderator variable (e.g., marketization level)
global w "Market"
// Clear previous regression results
eststo clear
// Baseline model
eststo: reghdfe $Y $D $X, absorb(pro year) vce(cluster pro)
// Model with interaction term
eststo: reghdfe $Y $D c.$D#c.$w $w $X, absorb(pro year) vce(cluster pro)
// Output results to RTF file
esttab using Moderating Effect.rtf, replace b(3) r2 ar2 star(* 0.1 ** 0.05 *** 0.01) nogap
```

*Figure 9.Endogeneity and Mechanism Analysis*

**Implementation:** Permutation tests (e.g., "permute") construct empirical distributions of estimates under the null. Counterfactual DID is simulated by randomizing treatment timing. The core S-DIDML estimator is re-run using LASSO, boosting, or neural networks under varying assumptions.

**Cautions:** Each robustness test should be accompanied by estimates, standard errors, and p-values. Report computational time and convergence where applicable.

### Stage 7: Heterogeneity Analysis (Group Effect Test)

**Purpose:** To explore whether estimated treatment effects are moderated by mechanisms (e.g., mediators) or vary across subgroups (e.g., urban-rural, resource-based cities).

```stata
clear
use data.dta, clear
// Restrict sample to resource-based cities
keep if inlist(resource_city,1)
// Re-define variables for subgroup analysis
global Y IGG
global X Edu Constru Urban Pass Fre Inv Inter Fis Unemp Size Consump Sci Cap ///
         Edu2 Constru2 Urban2 Pass2 Fre2 Inv2 Inter2 Fis2 Unemp2 Size2 ///
         Consump2 Sci2 Cap2 i.year i.id
global D Broadband
set seed 42
// Re-run DML for subgroup
ddml init partial, kfolds(5)
ddml E[D|X]: pystacked $D $X, type(reg) method(rf)
ddml E[Y|X]: pystacked $Y $X, type(reg) method(rf)
ddml crossfit
ddml estimate, robust
```

***Figure 10.Heterogeneity Analysis (Group Effect Test)***

**Implementation:** Mediation analysis can be conducted using a three-equation setup. Subgroup heterogeneity is analyzed by re-estimating the S-DIDML within strata. Interaction terms may be used to model moderators.

**Cautions:** Mediators must be theoretically justified. If endogenous, interpretation of causal paths must be conservative. Sufficient sample size is needed within subgroups for valid estimation.

## 5. Practical Roadmap for Applied Research

### 5.1 Workflow of S-DIDML in Practice

**Stage 1: Problem Formulation and Theoretical Foundations**

**Step 1: Defining Research Questions and Assessing Team Capabilities**

The research begins with the precise formulation of causal questions and empirical hypotheses. Concurrently, the research team’s capabilities are evaluated, including expertise in econometrics, data science, and domain-specific knowledge.

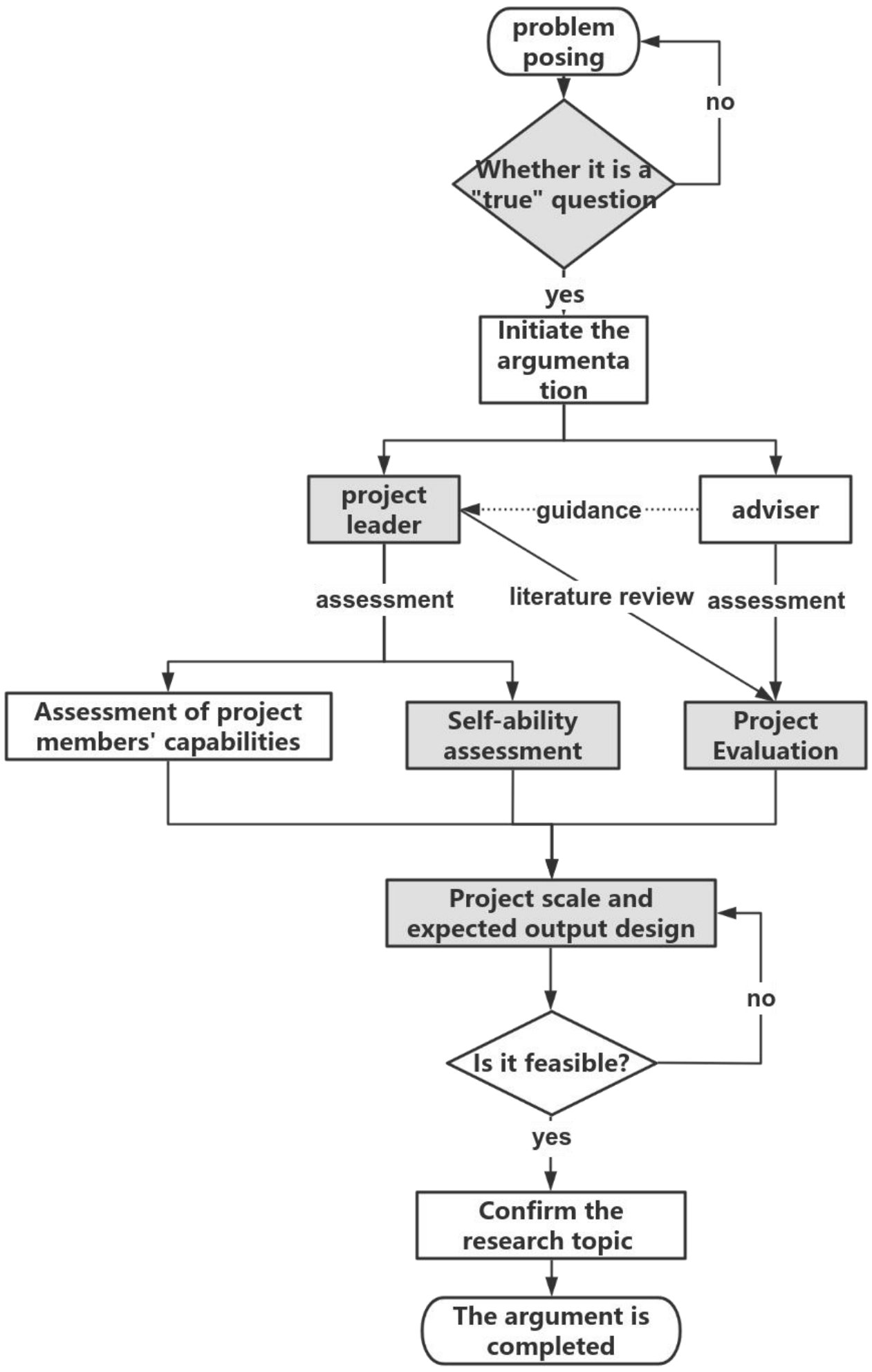

***Figure 11.Flow chart for determining research questions and assessing team capabilities***

**Step 2: Constructing Theoretical Framework and Panel Data Structure**

A theoretical causal model is articulated, often grounded in quasi-experimental logic. The panel structure (unit-time-treatment mapping) is defined, specifying whether the setup involves staggered treatment adoption, sharp assignment, or repeated interventions.

**Step 3: Data Collection and Preprocessing**

Panel or longitudinal data are acquired from credible sources (e.g., government databases, surveys, satellite data). Preprocessing includes dealing with missing values, removing outliers, aligning treatment timing, constructing dummies, and ensuring consistency across units and time.

**Stage 2: Variable Diagnostics and Model Initialization**

**Step 4: Descriptive Statistics and Correlation Analysis**

Summary statistics and correlation matrices are generated to understand the distribution, balance, and potential collinearity among covariates. Multicollinearity is assessed using tools such as the Variance Inflation Factor (VIF).

**Step 5: Decision on Dimension Reduction**

A diagnostic decision is made: If the number of covariates is large or multicollinearity is high, dimension reduction is warranted. If not, the original covariates are retained. Dimension reduction may include PCA, Factor Analysis, Cluster Analysis, or fsQCA.

**Step 6: Establishing Baseline Regressions**

Two benchmark models are estimated: A high-dimensional fixed effects regression using reghdfe to control for unobserved unit and time effects. A Double Machine Learning (DML) estimator that orthogonalizes treatment and outcome equations via cross-fitting and flexible learners.

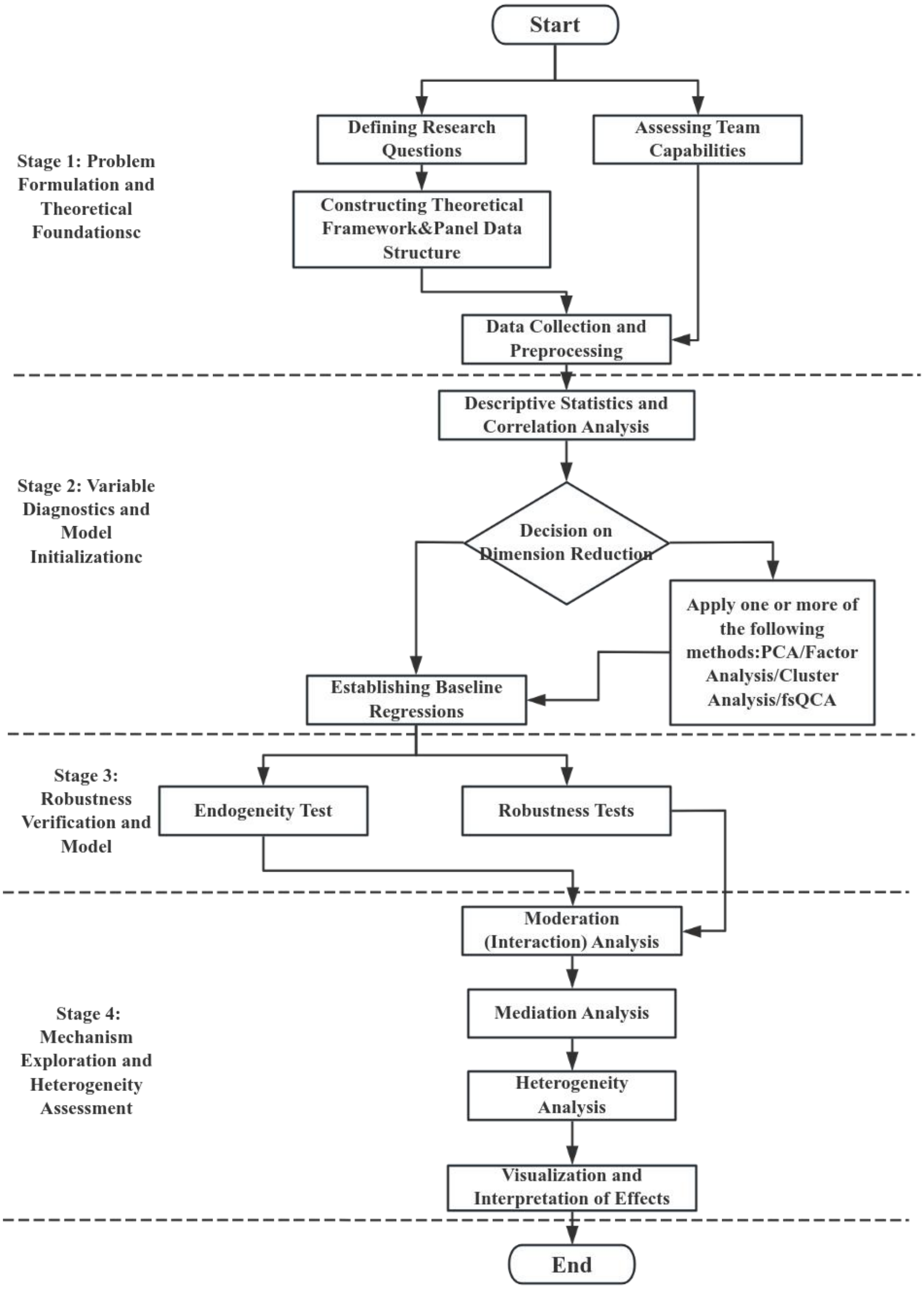

*Figure 12. Workflow of S-DIDML in Practice*

**Stage 3: Robustness Verification and Model Extension**

**Step 7: Robustness Tests**

A suite of empirical validity checks are conducted: Parallel trends test, placebo test, counterfactual test, winsorization test, timing interaction test, sample segmentation test, and method variation test.

**Step 8: Endogeneity Test**

If potential endogeneity of the treatment variable is suspected, the researcher applies instrumental variable-based DML (IV-DDML). This includes estimation of three nuisance functions: E[Y|X], E[D|X], and E[Z|X], where Z is a valid instrument.

**Stage 4: Mechanism Exploration and Heterogeneity Assessment**

**Step 9: Moderation (Interaction) Analysis**

Interaction terms are included to assess whether treatment effects vary across observed contextual factors. This allows identification of effect amplifiers or dampeners.

**Step 10: Mediation Analysis**

Mechanism channels are explored through multi-equation regression, identifying whether the treatment affects outcomes indirectly via mediating variables.

**Step 11: Heterogeneity Analysis**

Subgroup analyses are performed across strata. The S-DIDML estimator is re-estimated within each group to map treatment effect heterogeneity.

**Step 12: Visualization and Interpretation of Effects**

Estimated treatment effects are visualized using coefficient plots, confidence bands, or marginal effect graphs. Interpretations are aligned with policy context and theoretical expectations.

### 5.2 Model Selection and Tuning in Stata

#### 5.2.1 Model Selection

The selection of machine learning models under the S-DIDML framework should be a systematic decision-making process driven by data structure, theoretical assumptions, and causal inference objectives. Its core lies in enabling machine learning components to adapt to the complexity of covariates $X$ while preserving the structural rigor of difference-in-differences (DID). (Imbens & Rubin, 2015) Specifically, model selection must first address data characteristics: When $X$ contains high-dimensional interaction terms (e.g., the combined effects of education level, industrial structure, and digitalization on inclusive growth) or nonlinear relationships (such as the varying marginal effects of infrastructure investment on growth across economic development stages), random forests become preferred due to their ability to capture higher-order correlations. Their integration through multiple decision trees effectively resolves complex patterns in $X$. (Bach et al., 2024a)If $X$ has high dimensions but most variables theoretically exhibit linear relationships with outcomes (e.g., regional fiscal expenditure components and public service levels), LASSO proves more suitable through regularization-based sparsity processing. Its coefficient shrinkage capability for redundant variables simplifies models without losing critical information. Theoretical prior knowledge plays a crucial role in this process. For instance, when evaluating the effect of network infrastructure on economic growth, if the theory predicts "marketization moderates infrastructure's impact," the model must capture interaction effects. In such cases, random forests or gradient boosting outperform LASSO.(Ahrens et al., 2024a)Conversely, if the theory explicitly states "human capital is the core confounding factor," constraining the model's prioritization of human capital-related variables helps avoid interference from secondary variables. (Baiardi & Naghi, 2024a)Simultaneously, computational feasibility imposes practical constraints: While gradient boosting achieves improved fitting accuracy with large samples ($N$>10,000), its computational cost is 3-5 times higher than random forests, necessitating a trade-off between precision gains and time consumption. For small samples ($N$<500), overly complex models (e.g., random forests with depths exceeding 20) should be avoided to prevent overfitting caused by insufficient data support. It must be emphasized that model selection in machine learning should not prioritize "maximizing prediction accuracy" as a single objective, but rather focus on "assisting DID in handling the complexity of $X$". For instance, when $X$ exhibits multicollinearity, the model must possess noise reduction capabilities.(Kabata & Shintani, 2021) However, uncontrolled noise reduction may lead to excessive fitting of random fluctuations in $X$ thereby violating the parallel trend assumption —— of DID. Therefore, the ultimate validation criterion for model selection is whether the processed $X$ can be effectively integrated into the DID framework, yielding robust and interpretable causal estimates. (Collischon, 2022) Under this logic, S-DIDML's model selection approach transcends the limitations of mechanical application, truly becoming a reliable bridge connecting data complexity and rigorous structural integrity in high-dimensional policy evaluation.

#### 5.2.2 Model Recommendations for parameter adjustment

The parameter tuning of S-DIDML enhances machine learning components' adaptability to data characteristics through fine-tuning while maintaining the structural integrity of the DID framework. Its core principle lies in "targeted" —— tailored to $X$ complexity, sample size, and theoretical assumptions. For random forests, the number of trees (ntree) and maximum depth (maxdepth) should align with the dimensions of $X$ and sample size: At moderate dimensions ($p = 30$) and sample size ($N$=1000), ntree (500) and maxdepth (20) typically suffice to capture variable correlations. Blindly increasing the number of trees to 1000 would double computational time and risk noise-induced overfitting. In high-dimensional scenarios ($p = 100$) with large samples ($N$=5000), ntree (800) and maxdepth (30) enhance model recognition of complex patterns, though cross-validation must confirm their effectiveness. Whentree depth exceeds a threshold (e.g., 30), residual variance no longer decreases significantly, necessitating stopping deepening to avoid overfitting. The regularization parameter ($\lambda$) adjustment for LASSO requires theoretical guidance: For key variables with clear theories (e.g., "policy intensity" proxy variable in policy evaluation), reducing regularization weight prevents coefficient over-convergence; for noisy variables (e.g., secondary indicators with significant measurement errors), increasing penalty strength suppresses interference. The selection of cross-validation folds (kfolds) should be flexibly adjusted based on sample size: For small samples ($N$=300), $k = 3$ can avoid estimation fluctuations caused by insufficient sample size in single-fold scenarios. For large samples ($N$=10,000), $k = 8$ enhances bias correction effects, but the upper limit of folds should not exceed 1/5 of the sample size to prevent fold differences from becoming too small and undermining cross-validation effectiveness. Parameter tuning for gradient boosting requires more emphasis on "progressive optimization": Set the learningrate to 0.01 to prevent premature convergence to local optima, while iteration counts are controlled through early stopping rules.For example, in dynamic policy evaluation, if the mean squared error in the validation set shows no improvement after 50 consecutive iterations, iteration should be stopped to prevent overfitting. The effectiveness of parameter adjustments must be confirmed through "double checking": On one hand, residual variance in machine learning components should significantly decrease (indicating effective capture of valid information from $X$); on the other hand, core assumptions of DID (e.g., parallel trends) must still pass tests (indicating structural integrity). In summary, critical parameter tuning in S-DIDML is a collaborative process integrating data characteristics, theoretical assumptions, and computational feasibility, rather than mechanical parameter optimization. Success is marked by machine learning components that can both handle complex correlations in $X$ and provide stable support for the structural framework of DID. Once this balance is achieved, S-DIDML can output causal estimates with both reliability and interpretability in high-dimensional policy evaluation, which is the core advantage of S-DIDML compared to traditional DID or pure machine learning methods.

### 5.3 Diagnostics, Visualization, and Interpretation

#### 5.3.1 Descriptive statistics

**(1) Diagnostics**

This study included 2,820 observational samples spanning the period from 2010 to 2019 (with the mean value of the year variable being 2014.5). The data encompassed multiple core variables covering dependent variables, independent variables, and control variables, providing a robust informational foundation for subsequent analyses.

The mean value of the dependent variable IGG (Inclusive Green Growth Index) is 2.990508, with a standard deviation of 0.7973177 and a range from 0.9535196 to 7.199178. This indicates that while there are variations in inclusive green growth levels across samples, the overall distribution remains relatively concentrated. The other dependent variable EG has a mean of 1.726219, a standard deviation of 0.9972721, and fluctuates between 0.3159401 and 9.736856. Its relatively wide range suggests more pronounced differences in performance across different samples regarding this indicator.

The explanatory variable Broadband (broadband infrastructure) is a binary variable with a mean of 0.1882979, indicating that approximately 18.83% of the sample possesses this characteristic. Similarly, High-speed network (high-speed internet) is also a binary variable with a mean of 0.5078014, meaning about 50.78% of the sample has high-speed network access. Smartcity (smart city) has a mean of 0.3879433, while Bigdata (big data) has a mean of 0.1134752, suggesting that samples related to smart cities and big data account for approximately 38.79% and 11.35% respectively. Additionally, both year_2013 and year_2012 have means of 0.7, which may indicate their representativeness or special significance in the sample. The mean value of cityAdj is 0.8758865, indicating that most samples (approximately 87.59%) underwent urban adjustment-related processing.

Regarding control variables, resource_c~y has a mean of 0.4007092 and industrial_e has a mean of 0.3333333, likely indicating the distribution of resource-and industry-related characteristics within the sample. The class variable has a mean of 2.411348 with a standard deviation of 0.7633957, ranging from 1 to 3, suggesting it represents a categorical variable for sample classification. Other variables include Edu (mean 0.0345252), Constru (mean 0.9636387), Urban (mean 5.738196), Pass (mean 8.434764), Fre (mean 9.003546), Inv (mean 0.7791503), and Inter (mean 0.2118704). These variables characterize the sample through dimensions such as education, construction, urban development, transportation, freedom, investment, and interaction, each possessing distinct mean values and distribution ranges.

**(2) Visualization**

```
. *Descriptive statistics

.
. outreg2 using msxtj.doc, replace sum(log) keep($Y $D $X ) title(Decriptive statistics)
```

| Variable | Obs | Mean | Std. dev. | Min | Max |
|---|---|---|---|---|---|
| id | 2,820 | 141.5 | 81.42031 | 1 | 282 |
| pro | 2,820 | 15.1773 | 7.372695 | 1 | 31 |
| year | 2,820 | 2014.5 | 2.872791 | 2010 | 2019 |
| IGG | 2,820 | 2.990508 | .7973177 | .9535196 | 7.199178 |
| EG | 2,820 | 1.726219 | .9972721 | .3159401 | 9.736856 |
| ID | 2,820 | 3.09525 | 1.428762 | .2000862 | 9.275073 |
| WB | 2,820 | .6589715 | .6113799 | .0766691 | 7.3633 |
| PR | 2,820 | 6.481592 | 1.254982 | 1.6165 | 7.899693 |
| Broadband | 2,820 | .1882979 | .3910192 | 0 | 1 |
| High_speed | 2,820 | .5078014 | .5000278 | 0 | 1 |
| Smartcity | 2,820 | .3879433 | .4873679 | 0 | 1 |
| Bigdata | 2,820 | .1134752 | .317229 | 0 | 1 |
| year_2013 | 2,820 | .7 | .4583388 | 0 | 1 |
| year_2012 | 2,820 | .7 | .4583388 | 0 | 1 |
| city_adj | 2,820 | .8758865 | .3297694 | 0 | 1 |
| resource_c~y | 2,820 | .4007092 | .4901291 | 0 | 1 |
| industrial~e | 2,820 | .3333333 | .4714881 | 0 | 1 |
| class | 2,820 | 2.411348 | .7633957 | 1 | 3 |
| Edu | 2,820 | .0345252 | .0187168 | .0075343 | .1854739 |
| Constru | 2,820 | .9636387 | .5371972 | .1756098 | 5.154321 |
| Urban | 2,820 | 5.738196 | .9216112 | 1.619828 | 7.881657 |
| Pass | 2,820 | 8.434764 | 1.00364 | 2.197225 | 12.56569 |
| Fre | 2,820 | 9.003546 | .868714 | 5.361292 | 13.22529 |
| Inv | 2,820 | .7791503 | .2878309 | .0872265 | 2.243172 |
| Inter | 2,820 | .2118704 | .1835202 | .0034722 | 3.663528 |

*Figure 13.Descriptive statistics results visualization*

**(3) Interpretation**

Descriptive statistics provide essential insights into the dataset, including mean values, standard deviations, minimum and maximum values of variables. The mean and standard deviation of the dependent variables IGG and EG reflect the overall level and dispersion of inclusive green growth and related economic indicators. For explanatory variables, the means of broadband access, high-speed internet, smart city development, and big data demonstrate the adoption rate of these infrastructure technologies within the sample. Control variables offer additional descriptive dimensions to enhance sample analysis, effectively managing potential confounding factors in subsequent research.

However, descriptive statistics alone cannot provide sufficient insight into the relationships and underlying patterns among variables. Further statistical methods such as correlation analysis and regression analysis are required to explore how factors like network infrastructure specifically influence inclusive green growth mechanisms. Additionally, subsequent analyses should address potential issues including multicollinearity and heteroskedasticity between variables, with appropriate measures implemented to ensure the accuracy and reliability of research findings.

### 5.3.2 Correlation analysis

**(1) Diagnostics**

This correlation analysis covers PR, Broadband, Edu, Constru and other variables. By calculating the correlation coefficient between two variables, this study aims to explore the linear relationship between these variables and provide a basic reference for subsequent regression analysis.

Correlation between PR and other variables: The correlation coefficient between PR and Broadband is 0.043**, significant at the 1% level, indicating a strong positive correlation. This suggests that the development of broadband infrastructure may positively correlate with the indicators represented by PR. The correlation coefficient between PR and Education is-0.182**, also significant at the 1% level, suggesting a possible negative correlation between education level and PR. The correlation coefficient with Construction is 0.0170, which is not statistically significant, indicating a weak linear relationship between construction-related factors and PR.

Correlation between Broadband and Other Variables: The correlation coefficient between broadband and Education is 0.123***, indicating a significant positive correlation. This suggests that increased educational attainment may be accompanied by broader adoption of broadband infrastructure. The correlation coefficient with Construction is 0.194***, showing a significant positive correlation. This demonstrates that advancements in the construction sector might have a positive connection with the expansion of broadband infrastructure.

Correlation between other variables: For example, the correlation coefficient between Urban and Pass is 0.434***, indicating a significant positive correlation, which may reflect a strong positive linear relationship between urban-related factors and traffic-related indicators; the correlation coefficient between Fre and Pass is-0.132***, indicating a significant negative correlation, suggesting that free-related indicators and traffic-related indicators show an inverse trend of change.

**(2) Visualization**

| | PR | Broadband | Edu | Constru | Urban | Pass | Fre | Inv | Inter | Fis | Unemp | Size |
|---|---|---|---|---|---|---|---|---|---|---|---|---|
| PR | 1 | | | | | | | | | | | |
| Broadband | 0.043** | 1 | | | | | | | | | | |
| Edu | -0.182*** | -0.123*** | 1 | | | | | | | | | |
| Constru | 0.0170 | 0.194*** | 0.237*** | 1 | | | | | | | | |
| Urban | 0.500*** | 0.118*** | -0.326*** | -0.081*** | 1 | | | | | | | |
| Pass | 0.238*** | -0.048** | -0.170*** | -0.066*** | 0.434*** | 1 | | | | | | |
| Fre | 0.194*** | 0.221*** | -0.283*** | -0.0250 | 0.427*** | -0.132*** | 1 | | | | | |
| Inv | -0.106*** | -0.0190 | 0.439*** | 0.00600 | -0.251*** | 0.0250 | -0.0150 | 1 | | | | |
| Inter | 0.100*** | 0.321*** | -0.245*** | 0.293*** | 0.194*** | 0.389*** | 0.270*** | -0.00800 | 1 | | | |
| Fis | -0.249*** | -0.074*** | 0.799*** | 0.300*** | -0.434*** | 0.143*** | 0.063*** | -0.0140 | 0.379*** | 1 | | |
| Unemp | 0.00700 | -0.00300 | -0.0210 | -0.00700 | -0.0220 | 0.337*** | 0.076*** | -0.00600 | 0.267*** | 0.060*** | 1 | |
| Size | -0.052*** | 0.217*** | 0.156*** | 0.571*** | 0.00400 | 0.359*** | -0.128*** | -0.00100 | 0.325*** | -0.00300 | -0.00800 | 1 |
| Consump | 0.137*** | 0.152*** | 0.042** | 0.373*** | 0.235*** | -0.170*** | 0.814*** | -0.0160 | 0.191*** | -0.00100 | 0.126*** | 0.186*** |
| Sci | 0.149*** | 0.214*** | -0.037* | 0.196*** | 0.181*** | 0.221*** | 0.233*** | -0.00500 | 0.484*** | 0.235*** | 0.334*** | -0.289*** |
| Cap | 0.130*** | 0.159*** | -0.071*** | 0.138*** | 0.320*** | 0.217*** | -0.429*** | -0.0220 | 0.0160 | 0.234*** | 0.206*** | -0.160*** |
| Edu2 | -0.168*** | -0.074*** | 0.934*** | 0.243*** | -0.283*** | 0.00700 | -0.266*** | -0.0230 | 0.0180 | 0.053*** | 0.260*** | -0.243*** |
| Constru2 | 0.0110 | 0.145*** | 0.176*** | 0.932*** | -0.063*** | 0.199*** | -0.319*** | 0.00400 | 0.123*** | 0.133*** | -0.054*** | 0.440*** |
| Urban2 | 0.499*** | 0.134*** | -0.338*** | -0.071*** | 0.990*** | -0.158*** | 0.384*** | -0.00600 | 0.073*** | -0.0220 | 0.192*** | -0.065*** |
| Pass2 | 0.226*** | -0.040** | -0.173*** | -0.050*** | 0.423*** | 0.752*** | -0.054*** | 0.00900 | 0.164*** | 0.0290 | 0.0140 | 0.593*** |
| Fre2 | 0.187*** | 0.226*** | -0.274*** | -0.00800 | 0.417*** | -0.0240 | 0.833*** | -0.00600 | 0.223*** | 0.043** | -0.00300 | -0.00800 |
| Inv2 | -0.114*** | -0.0110 | 0.465*** | 0.073*** | -0.264*** | 0.0150 | -0.0110 | 0.999*** | -0.0160 | -0.0140 | 0.231*** | 0.167*** |
| Inter2 | 0.056*** | 0.114*** | -0.096*** | 0.121*** | 0.122*** | 0.244*** | 0.230*** | -0.00300 | 0.864*** | 0.258*** | -0.0160 | 0.00400 |
| Fis2 | -0.164*** | -0.034* | 0.477*** | 0.158*** | -0.235*** | 0.123*** | 0.073*** | -0.0140 | 0.369*** | 0.974*** | 0.045** | -0.00600 |
| Unemp2 | 0.0110 | -0.00900 | -0.0100 | -0.0110 | -0.0180 | 0.103*** | 0.053*** | -0.00200 | 0.097*** | 0.0160 | 0.999*** | -0.0110 |

| | Consump | Sci | Cap | Edu2 | Constru2 | Urban2 | Pass2 | Fre2 | Inv2 | Inter2 | Fis2 | Unemp2 |
|---|---|---|---|---|---|---|---|---|---|---|---|---|
| Consump | 1 | | | | | | | | | | | |
| Sci | -0.055*** | 1 | | | | | | | | | | |
| Cap | -0.0260 | 0.426*** | 1 | | | | | | | | | |
| Edu2 | -0.054*** | 0.422*** | 0.474*** | 1 | | | | | | | | |
| Constru2 | 0.066*** | -0.264*** | -0.144*** | -0.173*** | 1 | | | | | | | |
| Urban2 | 0.087*** | 0.140*** | 0.0170 | 0.101*** | -0.090*** | 1 | | | | | | |
| Pass2 | 0.118*** | -0.217*** | -0.146*** | -0.194*** | 0.181*** | -0.0140 | 1 | | | | | |
| Fre2 | -0.00800 | -0.0200 | -0.0200 | 0 | 0 | 0.00300 | -0.00400 | 1 | | | | |
| Inv2 | 0.366*** | -0.0130 | -0.0150 | 0.046** | 0.058*** | 0.099*** | 0.219*** | -0.00800 | 1 | | | |
| Inter2 | 0.234*** | 0.191*** | 0.0140 | 0.104*** | -0.0120 | 0.0190 | 0.046** | -0.0140 | 0.264*** | 1 | | |
| Fis2 | 0.077*** | 0.0300 | 0.0190 | 0.0260 | 0.042** | 0.066*** | 0.056*** | -0.00300 | 0.062*** | 0.00600 | 1 | |
| Unemp2 | 0.132*** | 0.330*** | 0.204*** | 0.256*** | -0.059*** | 0.199*** | 0.0150 | -0.00400 | 0.235*** | -0.0130 | 0.045** | 1 |

***Figure 14.*Correlation analysis visualization**

**（3）Interpretation**

The correlation analysis reveals significant relationships among multiple variables, providing crucial insights into their underlying connections. For instance, the positive correlation between PR and Broadband suggests that broadband infrastructure may positively impact sectors represented by PR, which aligns with our research on how network infrastructure influences inclusive green growth. Meanwhile, the negative correlation between PR and Education prompts further investigation into the mechanisms through which educational attainment affects these outcomes.

However, correlation analysis can only reveal linear relationships between variables and cannot directly establish causality. In subsequent analyses, we need to employ methods like regression models to investigate causal effects while controlling for potential confounding factors. Additionally, we must address issues such as multicollinearity. If certain variables show excessively high correlation coefficients, this may distort regression model estimates. To ensure the accuracy and reliability of subsequent analyses, appropriate measures like variable screening or principal component analysis should be implemented.

**5.3.3 Multicollinearity analysis**

**（1）Diagnostics**

This multicollinearity analysis was conducted using variance inflation factors (VIF) to evaluate the collinearity of multiple variables in the model. The results showed a Mean VIF of 157.61, significantly exceeding the threshold for minimal multicollinearity (typically <10). This indicates substantial multicollinearity issues within the model.

Cap and Cap2: The VIF value of Cap is up to 894.29, and the VIF value of Cap2 is 893.77, both far exceeding the standard of severe multiple collinearity (VIF>100), which may be due to the high collinearity caused by the quadratic transformation of Cap2.

Unemp and Unemp2: The VIF value of Unemp is 751.09, and the VIF value of Unemp2 is 750.67, which also has extreme multicollinearity. It is speculated that Unemp2 is the quadratic term of Unemp.

Fre and Fre2: The VIF value of Fre is 222.39, and the VIF value of Fre2 is 222.74, indicating serious collinearity, which may also be caused by the transformation of the second term.

Other variables: For instance, the VIF values for Pass and Pass2 are 108.32 and 108.89 respectively, while those for Urban and Urban2 stand at 60.09 and 60.27. Similarly, the VIF values for Consump and Consump2 are 23.55 and 22.03. All these variables exhibit varying degrees of severe collinearity, likely caused by their original variables and quadratic terms being jointly introduced into the model. In contrast, Broadband's VIF value of 1.28 is relatively low, indicating minimal collinearity with other variables.

**(2) Visualization**

```
. *Multicollinearity analysis

.
. reg $Y $D $X
```

| Source | SS | df | MS |
|---|---|---|---|
| Model | 1303.82755 | 27 | 48.2899093 |
| Residual | 3136.04404 | 2,792 | 1.12322494 |
| Total | 4439.87159 | 2,819 | 1.57498105 |

| | | |
|---|---|---|
| Number of obs | = | 2,820 |
| F(27, 2792) | = | 42.99 |
| Prob > F | = | 0.0000 |
| R-squared | = | 0.2937 |
| Adj R-squared | = | 0.2868 |
| Root MSE | = | 1.0598 |

| PR | Coefficient | Std. err. | t | P>\|t\| | [95% conf. | interval] |
|---|---|---|---|---|---|---|
| Broadband | -.1197687 | .0577066 | -2.08 | 0.038 | -.2329207 | -.0066167 |
| Edu | -8.291667 | 4.456806 | -1.86 | 0.063 | -17.03064 | .4473008 |
| Constru | .7652997 | .1354792 | 5.65 | 0.000 | .4996501 | 1.030949 |
| Urban | .2114312 | .1678959 | 1.26 | 0.208 | -.1177815 | .5406439 |
| Pass | .3717873 | .2069978 | 1.80 | 0.073 | -.034097 | .7776715 |
| Fre | .3168835 | .3426609 | 0.92 | 0.355 | -.3550109 | .9887778 |
| Inv | .0165323 | .2683503 | 0.06 | 0.951 | -.5096527 | .5427173 |
| Inter | .2158675 | .2334268 | 0.92 | 0.355 | -.2418391 | .6735741 |
| Fis | .4836796 | .6458035 | 0.75 | 0.454 | -.7826209 | 1.74998 |
| Unemp | -16.6464 | 4.463141 | -3.73 | 0.000 | -25.39779 | -7.895009 |
| Size | -.2130433 | .043999 | -4.84 | 0.000 | -.2993172 | -.1267694 |
| Consump | 2.164304 | .8842488 | 2.45 | 0.014 | .4304566 | 3.898152 |
| Sci | 67.81159 | 13.31516 | 5.09 | 0.000 | 41.70303 | 93.92014 |
| Cap | 6.163086 | 2.083429 | 2.96 | 0.003 | 2.07787 | 10.2483 |
| Edu2 | 20.3993 | 30.17013 | 0.68 | 0.499 | -38.75871 | 79.55731 |
| Constru2 | -.1212156 | .0328144 | -3.69 | 0.000 | -.1855585 | -.0568727 |
| Urban2 | .0386242 | .0157262 | 2.46 | 0.014 | .007788 | .0694604 |
| Pass2 | -.0197616 | .0121949 | -1.62 | 0.105 | -.0436736 | .0041504 |
| Fre2 | -.021159 | .0191238 | -1.11 | 0.269 | -.0586573 | .0163392 |
| Inv2 | .0510272 | .1392532 | 0.37 | 0.714 | -.2220224 | .3240769 |
| Inter2 | -.1150238 | .1114495 | -1.03 | 0.302 | -.3335555 | .103508 |
| Fis2 | -.4980874 | .2616122 | -1.90 | 0.057 | -1.01106 | .0148855 |
| Unemp2 | 2.585676 | .6854158 | 3.77 | 0.000 | 1.241703 | 3.929649 |
| Size2 | .0083456 | .0033746 | 2.47 | 0.013 | .0017287 | .0149626 |
| Consump2 | -2.379783 | 1.020469 | -2.33 | 0.020 | -4.380733 | -.3788332 |
| Sci2 | -1030.782 | 364.0157 | -2.83 | 0.005 | -1744.549 | -317.0146 |
| Cap2 | -.42561 | .143036 | -2.98 | 0.003 | -.7060769 | -.1451431 |
| _cons | -21.67399 | 7.592398 | -2.85 | 0.004 | -36.56127 | -6.786709 |

***Figure15.*Multicollinearity analysis visualization (1)**

```
. estat vif
```

| Variable | VIF | 1/VIF |
|---|---|---|
| Cap | 894.29 | 0.001118 |
| Cap2 | 893.77 | 0.001119 |
| Unemp | 751.09 | 0.001331 |
| Unemp2 | 750.67 | 0.001332 |
| Fre2 | 222.74 | 0.004490 |
| Fre | 222.39 | 0.004497 |
| Pass2 | 108.89 | 0.009184 |
| Pass | 108.32 | 0.009232 |
| Urban2 | 60.27 | 0.016593 |
| Urban | 60.09 | 0.016642 |
| Consump | 23.55 | 0.042463 |
| Consump2 | 22.03 | 0.045392 |
| Fis | 18.95 | 0.052774 |
| Edu | 17.46 | 0.057261 |
| Inv2 | 15.01 | 0.066629 |
| Inv | 14.97 | 0.066787 |
| Constru | 13.29 | 0.075224 |
| Edu2 | 12.27 | 0.081506 |
| Constru2 | 10.25 | 0.097587 |
| Fis2 | 8.35 | 0.119744 |
| Size | 7.56 | 0.132252 |
| Size2 | 4.90 | 0.204064 |
| Inter | 4.61 | 0.217121 |
| Sci | 3.09 | 0.323935 |
| Inter2 | 2.79 | 0.357897 |
| Sci2 | 2.48 | 0.403831 |
| Broadband | 1.28 | 0.782571 |
| Mean VIF | 157.61 | |

***Figure 16.*Multicollinearity analysis visualization (2)**

**(3) Interpretation**

The multicollinearity analysis reveals significant multicollinearity issues in the model, primarily caused by introducing quadratic terms for multiple variables. This high degree of collinearity may lead to unstable estimation of regression coefficients and increased standard errors, thereby affecting the results of hypothesis testing and the interpretability of the model. For instance, the extreme VIF values of Cap and Cap2 significantly reduce the precision of their coefficient estimates, making it difficult to accurately assess their true impact on the dependent variable.

In subsequent analysis, we need to implement appropriate measures to mitigate multicollinearity issues. For instance, we may consider performing dimensionality reduction on highly collinear variables through methods like Principal Component Analysis (PCA) to extract core information, or remove one of the highly correlated variables to reduce collinearity effects.

Simultaneously, it is crucial to reassess the model specification—consider whether it's necessary to introduce both the original variables and their binomial transformations, or adopt alternative approaches to capture nonlinear relationships between variables. This ensures the accuracy and reliability of analytical results while more effectively exploring the mechanisms by which factors influence the dependent variable.

### 5.3.4 Principal component analysis

**(1) Diagnostics**

Principal component analysis (PCA) is conducted using standardized variables to address multicollinearity through dimensionality reduction. The results include principal components (eigenvalues), Kaiser-Meyer-Olkin (KMO) test, as well as scatter plots and component loadings, providing a comprehensive evaluation of the method's applicability and effectiveness from multiple perspectives.

The principal component (feature vector) table reveals distinct variable loadings across components. For instance, standard education (stdEdu) shows a-0.2979 loading on Component 1 and a 0.1511 loading on Component 2, while standard construction (stdConstru) exhibits-0.0628 on Component 1 versus 0.3582 on Component 2. These loadings indicate the correlation between variables and principal components, where higher absolute values signify stronger representativeness of variables in specific components. By analyzing the distribution of variable loadings across components, researchers can preliminarily identify the underlying factors represented by these principal components.

The Kaiser-Meyer-Olkin (KMO) test is employed to evaluate partial correlations between variables and determine suitability for principal component analysis. The results show that most variables have KMO values between 0.45 and 0.75, with an overall KMO value of 0.6320. Generally, KMO values above 0.6 indicate sufficient data quality for PCA, suggesting this dataset may be suitable for dimensionality reduction through PCA. However, certain variables such as stdUnemp (0.4580), stdUnemp2 (0.4576), and stdSci2 (0.4825) exhibit relatively low KMO values, which could potentially affect the effectiveness of PCA analysis.

The scree plot illustrates how eigenvalues of principal components evolve with their number. As shown, the eigenvalues of early principal components decrease rapidly before stabilizing. Typically, we select components with eigenvalues greater than 1 or determine the number of principal components based on inflection points in the plot. In this study, the initial principal components exhibit high eigenvalues that account for significant variance, while subsequent components show decreasing eigenvalues and reduced explanatory power.

A component loading plot visually illustrates how variables are distributed across two principal components. By analyzing the plot, we can identify which variables exhibit high loadings on the same component, indicating they may collectively reflect a latent factor. For instance, when certain variables show concentrated loading on both Component1 and Component 2, this suggests their strong correlation with these two principal components.

**(2) Visualization**

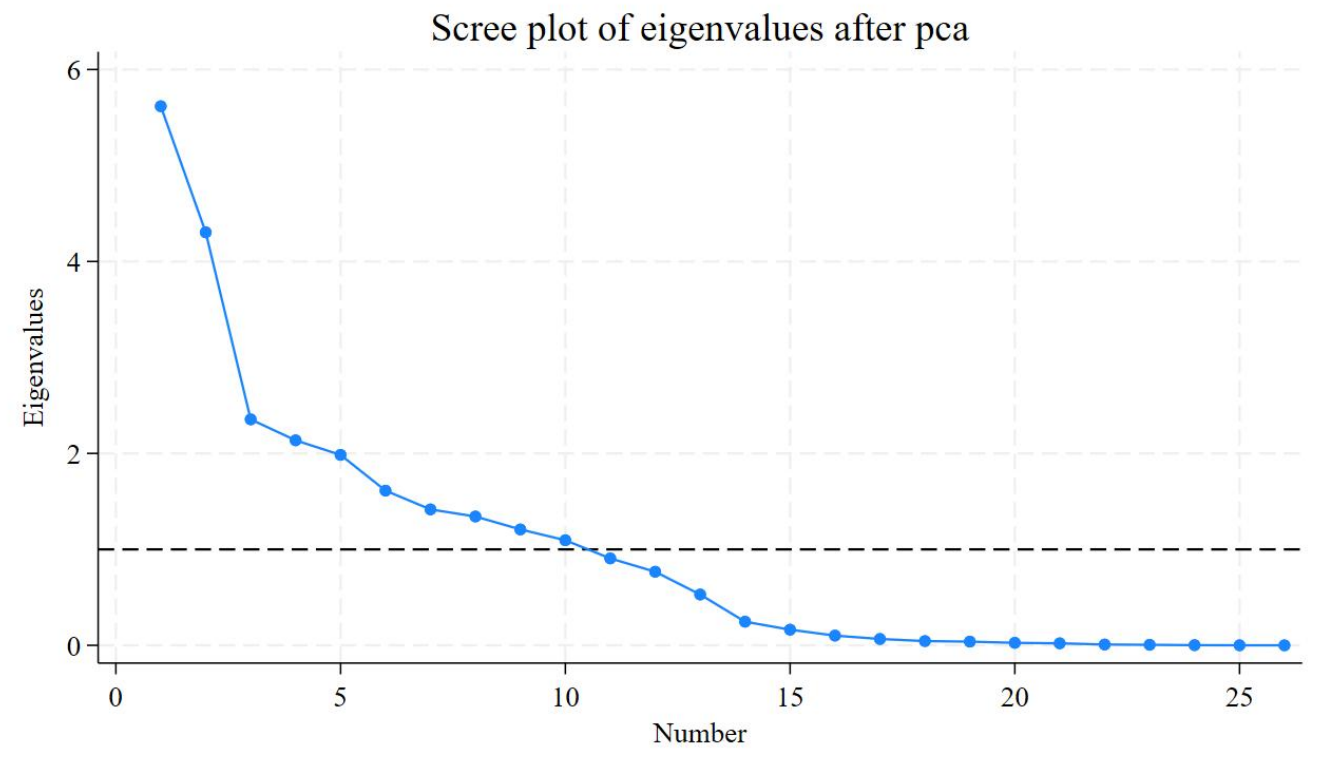

*Figure 17.Scree plot*

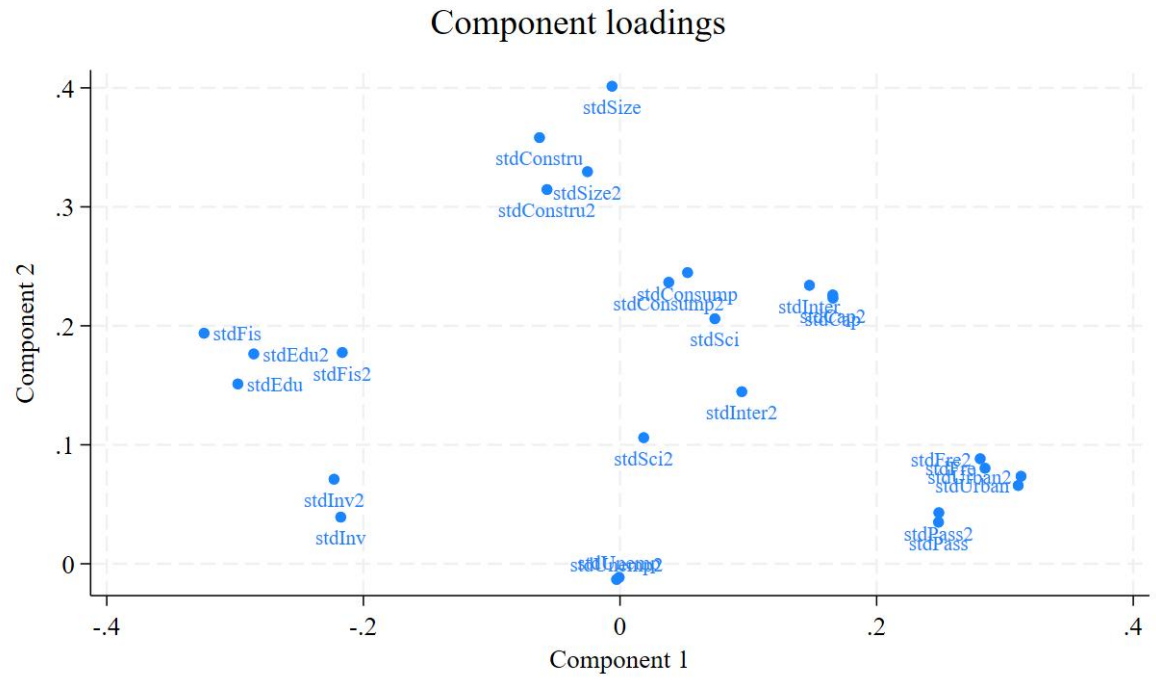

*Figure 18.Component load diagram*

Principal components (eigenvectors)

| Variable | Comp1 | Comp2 | Comp3 | Comp4 | Comp5 | Comp6 | Comp7 | Comp8 | Comp9 | Comp10 | Unexplained |
|---|---|---|---|---|---|---|---|---|---|---|---|
| stdEdu | -0.2979 | 0.1511 | 0.2633 | -0.0545 | 0.0424 | -0.0684 | -0.0947 | 0.1370 | 0.0634 | 0.1055 | .1676 |
| stdConstru | -0.0628 | 0.3582 | -0.1754 | -0.0926 | 0.0119 | -0.0130 | -0.0888 | -0.2883 | -0.1045 | 0.2916 | .1053 |
| stdUrban | 0.3103 | 0.0658 | 0.1185 | -0.0963 | 0.0262 | 0.0041 | 0.0894 | 0.3462 | -0.1658 | 0.1727 | .1481 |
| stdPass | 0.2482 | 0.0350 | 0.3477 | -0.1341 | 0.0707 | 0.0291 | -0.2465 | -0.1609 | -0.0202 | 0.2406 | .1295 |
| stdFre | 0.2845 | 0.0803 | 0.2122 | -0.0748 | 0.0763 | 0.0328 | 0.0208 | -0.1853 | 0.3546 | -0.2761 | .1044 |
| stdInv | -0.2176 | 0.0392 | 0.3156 | 0.0186 | 0.0496 | 0.0891 | 0.5275 | -0.1438 | 0.0172 | 0.0874 | .04369 |
| stdInter | 0.1476 | 0.2341 | -0.2263 | 0.2795 | -0.0887 | 0.0261 | 0.0984 | 0.0862 | 0.3680 | 0.1923 | .1098 |
| stdFis | -0.3242 | 0.1937 | 0.1562 | -0.0120 | 0.0160 | 0.0160 | -0.2248 | 0.1780 | 0.1285 | -0.0437 | .05316 |
| stdUnemp | -0.0007 | -0.0114 | -0.0815 | 0.2024 | 0.6712 | 0.0095 | -0.0208 | 0.0158 | -0.0231 | 0.0100 | .0004565 |
| stdSize | -0.0062 | 0.4014 | -0.0954 | -0.0060 | -0.0003 | -0.0813 | 0.0007 | -0.1635 | -0.0065 | -0.2523 | .1686 |
| stdConsump | 0.0526 | 0.2447 | -0.1566 | -0.4498 | 0.1231 | 0.0640 | 0.1815 | 0.2277 | 0.0108 | -0.0488 | .08105 |
| stdSci | 0.0739 | 0.2059 | 0.0562 | 0.2166 | -0.0699 | 0.5697 | -0.0529 | 0.0519 | -0.1266 | -0.0669 | .1144 |
| stdCap | 0.1660 | 0.2233 | 0.1432 | 0.3178 | -0.0725 | -0.3204 | 0.1147 | 0.0883 | -0.2528 | -0.1047 | .07241 |
| stdEdu2 | -0.2854 | 0.1764 | 0.2781 | -0.0011 | 0.0303 | -0.0769 | -0.1451 | 0.1791 | 0.0775 | 0.0795 | .1278 |
| stdConstru2 | -0.0569 | 0.3145 | -0.1649 | -0.0536 | -0.0009 | -0.0145 | -0.1126 | -0.3477 | -0.1894 | 0.3569 | .1225 |
| stdUrban2 | 0.3125 | 0.0736 | 0.1148 | -0.0859 | 0.0219 | 0.0043 | 0.0863 | 0.3471 | -0.1587 | 0.1716 | .1453 |
| stdPass2 | 0.2485 | 0.0429 | 0.3438 | -0.1271 | 0.0691 | 0.0253 | -0.2581 | -0.1735 | -0.0116 | 0.2316 | .1283 |
| stdFre2 | 0.2807 | 0.0883 | 0.2139 | -0.0738 | 0.0759 | 0.0329 | 0.0082 | -0.1903 | 0.3590 | -0.2748 | .1041 |
| stdInv2 | -0.2229 | 0.0711 | 0.3003 | 0.0138 | 0.0479 | 0.0877 | 0.5180 | -0.1490 | 0.0282 | 0.1299 | .03998 |
| stdInter2 | 0.0949 | 0.1446 | -0.1549 | 0.2745 | -0.0938 | 0.0407 | 0.0741 | 0.1615 | 0.5203 | 0.3590 | .1107 |
| stdFis2 | -0.2165 | 0.1776 | 0.1489 | 0.0442 | 0.0010 | -0.0070 | -0.3070 | 0.3080 | 0.1200 | -0.1490 | .242 |
| stdUnemp2 | -0.0027 | -0.0132 | -0.0746 | 0.1987 | 0.6730 | 0.0093 | -0.0209 | 0.0206 | -0.0270 | 0.0155 | .0004717 |
| stdSize2 | -0.0254 | 0.3295 | -0.0902 | -0.0038 | 0.0006 | -0.1077 | -0.0194 | -0.1670 | -0.0381 | -0.3401 | .3247 |
| stdConsump2 | 0.0380 | 0.2366 | -0.1730 | -0.4505 | 0.1213 | 0.0526 | 0.1882 | 0.2220 | 0.0093 | -0.0682 | .0917 |
| stdSci2 | 0.0184 | 0.1060 | 0.0275 | 0.1786 | -0.0629 | 0.6420 | -0.0398 | 0.0283 | -0.2267 | -0.1432 | .1197 |
| stdCap2 | 0.1658 | 0.2259 | 0.1356 | 0.3195 | -0.0742 | -0.3201 | 0.1116 | 0.0876 | -0.2508 | -0.1025 | .07304 |

***Figure 19.Principal component analysis（1）***

. estat kmo

Kaiser-Meyer-Olkin measure of sampling adequacy

| Variable | kmo |
|---|---|
| stdEdu | 0.6349 |
| stdConstru | 0.6593 |
| stdUrban | 0.7105 |
| stdPass | 0.6300 |
| stdFre | 0.6534 |
| stdInv | 0.6063 |
| stdInter | 0.6433 |
| stdFis | 0.7139 |
| stdUnemp | 0.4580 |
| stdSize | 0.6987 |
| stdConsump | 0.5879 |
| stdSci | 0.6140 |
| stdCap | 0.6131 |
| stdEdu2 | 0.7376 |
| stdConstru2 | 0.6265 |
| stdUrban2 | 0.7149 |
| stdPass2 | 0.6286 |
| stdFre2 | 0.6480 |
| stdInv2 | 0.6199 |
| stdInter2 | 0.5485 |
| stdFis2 | 0.5384 |
| stdUnemp2 | 0.4576 |
| stdSize2 | 0.6353 |
| stdConsump2 | 0.5892 |
| stdSci2 | 0.4825 |
| stdCap2 | 0.6164 |
| Overall | 0.6320 |

***Figure 20.Principal component analysis（2）***

**（3）Interpretation**

The principal component analysis (PCA) results indicate that the data in this study are suitable for PCA analysis. Through dimensionality reduction, multiple correlated variables can be transformed into a few uncorrelated principal components, thereby reducing the impact of multicollinearity. The PCA table, KMO test, scatter plot, and component loading plot provide comprehensive insights into the PCA process from different perspectives.

However, principal component analysis (PCA) has certain limitations. For example, the interpretation of PCA components may lack intuitive clarity, requiring professional knowledge to understand the underlying factors they represent. Additionally, some variables exhibit low KMO values, which could potentially affect the effectiveness of PCA analysis and should be considered in subsequent analyses.

In subsequent research, we can select an appropriate number of principal components based on the results of principal component analysis to introduce them as new variables into the regression model. This approach helps mitigate multicollinearity issues while enhancing model accuracy and reliability. Additionally, it is essential to validate the effectiveness of principal component analysis through comparative evaluations of metrics such as model fit indices and coefficient significance before and after the analysis, thereby assessing its role in resolving multicollinearity problems.

#### 5.3.5 High-dimensional fixed effects benchmark regression

**(1) Diagnostics**

This high-dimensional fixed-effects benchmark regression model, with PR as the dependent variable, was estimated using the reghdfe command. It incorporates high-dimensional fixed effects for both individual (id) and time (year) variables, while performing cluster robust standard error adjustments at the individual level. The model contains 2,820 observed samples across 282 clusters (individuals). The F-statistic reaches 3.22 with a corresponding P-value of 0.0000, indicating overall statistical significance. The R-square value stands at 0.7550, with an adjusted Adj R-square of 0.7240, demonstrating strong explanatory power for the dependent variable PR. The Within R-square of 0.0607 reflects the degree of variation explained within groups. The Root Mean Squared Error (RMSE) measures residual variance at 0.6593, indicating the average level of model residuals.

Broadband: The coefficient was-0.0477265 with a standard error of 0.0927761. The t-value was-0.51 and the P-value was 0.607, indicating no significant effect. The 95% confidence interval ranged from [-0.2303509 to 0.1348979], which included the value of 0, suggesting that broadband infrastructure has no statistically significant impact on PR.

Edu: The coefficient is 3.528313, the standard error is 8.534017, the t-value is 0.41, and the P-value is 0.680, which is not significant. This indicates that the influence of education-related factors on PR has not passed the significance test in the current model.

Constru: The coefficient was 0.2966332, the standard error was 0.2141815, the t-value was 1.38, and the P value was 0.167, which was not significant. The influence of construct-related factors on PR was statistically insignificant.

Other variables: For example, the Urban coefficient was 5.07941 with a t-value of 1.13 and P value of 0.259 (not significant); the Pass coefficient was 0.2825403 with a t-value of 1.09 and P value of 0.277 (not significant); and the Sci coefficient was 13.1655 with a t-value of 0.54 and P value of 0.587 (not significant).

**(2) Visualization**

| variable name | (1) PR | (2) PR | variable name | (1) PR | (2) PR |
|---|---|---|---|---|---|
| Broadband | -0.020 | -0.048 | Edu2 | | 79.054 |
| | (-0.21) | (-0.51) | | | (1.55) |
| Edu | | 3.528 | Constru2 | | -0.061 |
| | | (0.41) | | | (-1.50) |
| Constru | | 0.297 | Urban2 | | -0.375 |
| | | (1.38) | | | (-1.06) |
| Urban | | 5.079 | Pass2 | | -0.015 |
| | | (1.13) | | | (-0.95) |
| Pass | | 0.283 | Fre2 | | -0.035 |
| | | (1.09) | | | (-1.32) |
| Fre | | 0.725 | Inv2 | | -0.429 |
| | | (1.46) | | | (-1.58) |
| Inv | | 0.698 | Inter2 | | 0.019 |
| | | (1.33) | | | (0.26) |
| Inter | | -0.002 | Fis2 | | -0.016 |
| | | (-0.01) | | | (-0.05) |
| Fis | | -0.314 | Unemp2 | | 1.004 |
| | | (-0.36) | | | (1.17) |
| Unemp | | -6.528 | Size2 | | -0.005 |
| | | (-1.17) | | | (-1.25) |
| Size | | 0.100 | Consump2 | | -0.794 |
| | | (1.25) | | | (-0.64) |
| Consump | | 1.239 | Sci2 | | -56.040 |
| | | (1.08) | | | (-0.13) |
| Sci | | 13.165 | Cap2 | | -0.868 |
| | | (0.54) | | | (-1.47) |
| Cap | | 13.618 | _cons | 6.485*** | -69.115** |
| | | (1.58) | | (367.30) | (-2.19) |
| *N* | 2820 | 2820 | *N* | 2820 | 2820 |
| $R^2$ | 0.739 | 0.755 | $R^2$ | 0.739 | 0.755 |
| adj. $R^2$ | 0.709 | 0.724 | adj. $R^2$ | 0.709 | 0.724 |

*t* statistics in parentheses
* $p < 0.1$, ** $p < 0.05$, *** $p < 0.01$

*Figure 21.High-dimensional fixed effects benchmark regression（1）*

```
. eststo:reghdfe $Y $D $X,absorb(id year) vce(cluster id)
(MWFE estimator converged in 2 iterations)

HDFE Linear regression                            Number of obs   =      2,820
Absorbing 2 HDFE groups                           F(  27,    281) =       3.22
Statistics robust to heteroskedasticity           Prob > F        =     0.0000
                                                  R-squared       =     0.7550
                                                  Adj R-squared   =     0.7240
                                                  Within R-sq.    =     0.0607
Number of clusters (id)      =        282         Root MSE        =     0.6593

                                   (Std. err. adjusted for 282 clusters in id)
```

| PR | Coefficient | Robust std. err. | t | P>\|t\| | [95% conf. | interval] |
|---|---|---|---|---|---|---|
| Broadband | -.0477265 | .0927761 | -0.51 | 0.607 | -.2303509 | .1348979 |
| Edu | 3.528313 | 8.534017 | 0.41 | 0.680 | -13.27041 | 20.32703 |
| Constru | .2966332 | .2141815 | 1.38 | 0.167 | -.1249706 | .718237 |
| Urban | 5.07941 | 4.486619 | 1.13 | 0.259 | -3.75224 | 13.91106 |
| Pass | .2825403 | .2594379 | 1.09 | 0.277 | -.2281481 | .7932288 |
| Fre | .7253252 | .49591 | 1.46 | 0.145 | -.2508448 | 1.701495 |
| Inv | .6979102 | .5237398 | 1.33 | 0.184 | -.3330413 | 1.728862 |
| Inter | -.0016781 | .2776232 | -0.01 | 0.995 | -.5481633 | .5448072 |
| Fis | -.313698 | .8604713 | -0.36 | 0.716 | -2.007486 | 1.38009 |
| Unemp | -6.528476 | 5.599364 | -1.17 | 0.245 | -17.5505 | 4.493547 |
| Size | .0996685 | .0797055 | 1.25 | 0.212 | -.0572273 | .2565642 |
| Consump | 1.238931 | 1.144111 | 1.08 | 0.280 | -1.013187 | 3.491048 |
| Sci | 13.1655 | 24.21982 | 0.54 | 0.587 | -34.50982 | 60.84081 |
| Cap | 13.61811 | 8.593725 | 1.58 | 0.114 | -3.298137 | 30.53436 |
| Edu2 | 79.05402 | 50.90493 | 1.55 | 0.122 | -21.14939 | 179.2574 |
| Constru2 | -.0614194 | .0409246 | -1.50 | 0.135 | -.1419771 | .0191383 |
| Urban2 | -.3747841 | .3531642 | -1.06 | 0.290 | -1.069967 | .3203991 |
| Pass2 | -.0146481 | .0153881 | -0.95 | 0.342 | -.0449388 | .0156425 |
| Fre2 | -.0353214 | .0266603 | -1.32 | 0.186 | -.0878007 | .0171579 |
| Inv2 | -.4294679 | .2725741 | -1.58 | 0.116 | -.9660142 | .1070783 |
| Inter2 | .0190202 | .0743127 | 0.26 | 0.798 | -.12726 | .1653005 |
| Fis2 | -.0160812 | .2981546 | -0.05 | 0.957 | -.6029814 | .5708189 |
| Unemp2 | 1.004155 | .8601789 | 1.17 | 0.244 | -.6890568 | 2.697368 |
| Size2 | -.0048233 | .0038494 | -1.25 | 0.211 | -.0124005 | .002754 |
| Consump2 | -.7939848 | 1.249657 | -0.64 | 0.526 | -3.253862 | 1.665892 |
| Sci2 | -56.04036 | 444.5537 | -0.13 | 0.900 | -931.1186 | 819.0379 |
| Cap2 | -.8680856 | .5908698 | -1.47 | 0.143 | -2.031179 | .2950074 |
| _cons | -69.11502 | 31.51961 | -2.19 | 0.029 | -131.1596 | -7.070486 |

***Figure 22.High-dimensional fixed effects benchmark regression （2）***

**（3）Interpretation**

According to the results of high-dimensional fixed effects benchmark regression, the model has a good explanatory power on the whole, but the influence of multiple explanatory variables on the dependent variable PR is not statistically significant. This may be caused by the following reasons:

Variable measurement error: Some variables may not be able to accurately measure their corresponding concepts, resulting in inaccurate estimates.

Model setting problem: The current model may not have fully considered the nonlinear relationship between variables, interaction effect or other complex influence mechanism, and the model setting needs to be further improved.

Multiple collinearity problem: although principal component analysis and other treatments have been carried out, there may still be a certain degree of multiple collinearity, which affects the estimation accuracy and significance of coefficients.

In subsequent analysis, we need to further investigate the underlying reasons behind these insignificant results. For instance, we could improve variable measurement methods, add or adjust control variables, and consider nonlinear relationships or interaction effects to enhance model accuracy and interpretability. Additionally, robustness checks such as modifying model specifications or employing alternative estimation methods can be conducted to validate the reliability of the findings.

### 5.3.6 Dual-machine learning benchmark regression

**(1) Diagnostics**

This dual-machine learning benchmark regression model employs partial cross-validation (crossfit folds k=5) and resampling (resamples r=1) methods, with PR as the dependent variable. The Y1_pystacked learner estimates the conditional expectation of the dependent variable, while the D1_pystacked learner calculates the conditional expectation of the explanatory variable Broadband. The model incorporates 2,820 observed samples, with robust standard errors used for estimation.

Broadband: The coefficient was-0.0085482 with a standard error of 0.0521965. The z-value was-0.16 and the P-value was 0.870, indicating no significant effect. The 95% confidence interval ranged from [-0.1108515 to 0.093755], which included the value of 0, suggesting that broadband infrastructure has no statistically significant impact on PR.

Cons: The coefficient is 0.0297345, the standard error is 0.0144662, the z-value is 2.06, and the P-value is 0.040, which is significant at the 5% level. The 95% confidence interval is [0.0013814,0.0580877], excluding 0, indicating that the intercept term of the model is statistically significant.

**(2) Visualization**

```
Model:                     partial, crossfit folds k=5, resamples r=1
Mata global (mname):       m0
Dependent variable (Y):    PR
 PR learners:              Y1_pystacked
D equations (1):           Broadband
 Broadband learners:       D1_pystacked

DDML estimation results:
spec  r     Y learner     D learner         b        SE
   1  1  Y1_pystacked  D1_pystacked    -0.009   (0.052)

DDML model
y-E[y|X]  = y-Y1_pystacked_1                       Number of obs   =      2820
D-E[D|X]  = D-D1_pystacked_1
```

| PR | Coefficient | Robust std. err. | z | P>\|z\| | [95% conf. | interval] |
|---|---|---|---|---|---|---|
| Broadband | -.0085482 | .0521965 | -0.16 | 0.870 | -.1108515 | .093755 |
| _cons | .0297345 | .0144662 | 2.06 | 0.040 | .0013814 | .0580877 |

*Figure 23.Dual-machine learning benchmark regression*

**(3) Interpretation**

According to the regression results of the dual machine learning benchmark, the impact of Broadband on PR is not statistically significant, which is consistent with the previous high-dimensional fixed effect benchmark regression results. This may be caused by the following reasons:

Variable selection problem: The current selected learner may not have fully captured the complex relationship between variables, and other learners or parameters of the learner need to be tried.

Sample size problem: Although the sample size is large, it may not be sufficient to accurately estimate the impact of Broadband on PR for the current model setup.

Model specification questions: The dual machine learning model may require further adjustments, such as adding more control variables, considering interaction effects or other complex mechanisms of influence.

In the follow-up analysis, we can try the following improvement measures:

Change the learner: Try using other types of learners, such as random forest, gradient boosting tree, etc., to capture nonlinear relationships between variables.

Adjust model parameters: Adjust the number of cross-validation, resampling times and other parameters to improve the estimation accuracy of the model.

Add control variables: Consider adding more control variables to reduce the impact of omitted variable bias.

Through these improvements, we hope to more accurately estimate the impact of Broadband on PR and improve the interpretability and reliability of the model.

**5.3.7 Parallel trend test**

**(1) Diagnostics**

This parallel trend test graph illustrates the evolution of policy effects (Policy effect) at different time points before and after policy implementation. The horizontal axis represents relative policy implementation time (Parallel Trend Test of Policies), with the zero point marking the policy implementation date. Negative values indicate pre-implementation periods, while positive values denote post-implementation periods. The vertical axis displays estimated policy effects, where red dashed lines represent implementation time points, blue dots indicate estimated effect values at each time point, and error bars denote confidence intervals for these estimates.

Pre-policy implementation period (-3, -2, -1 phases): During the initial-3 and-2 phases prior to policy implementation, the estimated policy effects remained at approximately 0.2 with narrow confidence intervals, indicating parallel trends in outcome variables between treatment and control groups. By the-1 phase, the effect estimate rose to around 0.4 with expanded confidence intervals, though still not reaching zero. This suggests potential anticipatory effects of the policy that require further verification.

After policy implementation (Phase 1, 2, 3, and 4): In Phase 1 after policy implementation, the estimated effect of the policy decreased to about 0.1, and the estimated values in Phase 2 and later were close to 0, with the confidence interval containing 0, indicating that the trend of the outcome variables in the treatment group and the control group tended to be consistent after policy implementation, and the parallel trend hypothesis was established after policy implementation.

**(2) Visualization**

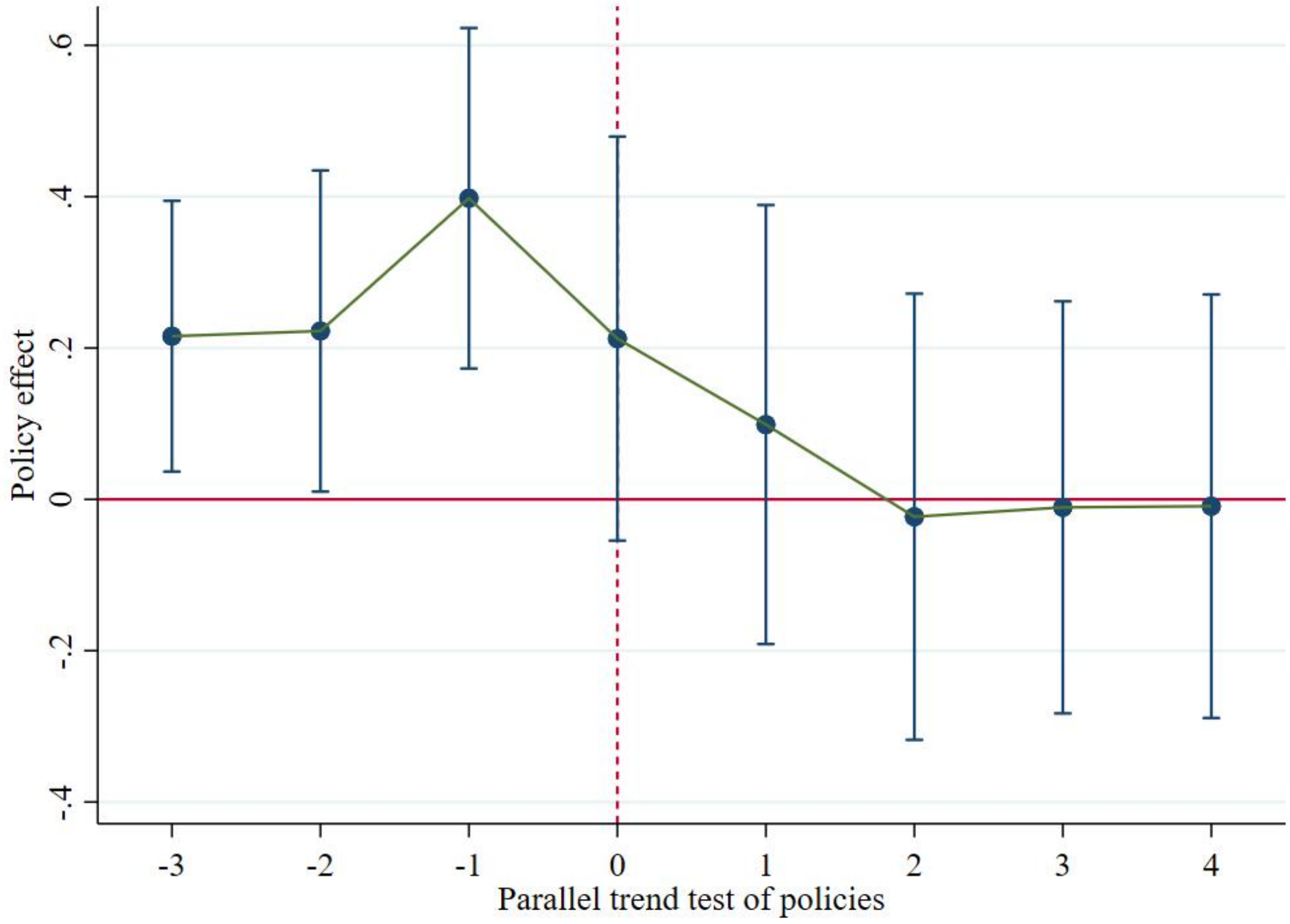

***Figure 24.Parallel trend test***

**（3）Interpretation**

The parallel trend test results show that in the pre-policy implementation periods (-3 and-2), both the treatment group and control group exhibited relatively consistent trend patterns in outcome variables, meeting the basic requirements of the parallel trend hypothesis. However, during the pre-policy implementation period (-1), the estimated policy effect showed a significant increase, which may indicate an anticipatory policy effect. This suggests that the treatment group had already begun adjusting their behavior before the policy implementation, potentially affecting the estimation results of the difference-in-differences model.

After the policy implementation, the trend of outcome variables in both the treatment group and control group converged, confirming the parallel trend hypothesis post-policy implementation. This validates the validity of the difference-in-differences model. However, it should be noted that pre-policy implementation expectations may introduce bias into estimated policy effects. Further robustness checks, such as placebo tests and sample period adjustments, are required to confirm the reliability of the results.

In the follow-up analysis, we can consider the following improvements:

Control for expected effects: A time dummy variable before policy implementation is added to the model to control for the effect of expected effects.

Shorten the sample period: shorten the sample period to a shorter time range before the implementation of the policy to reduce the impact of the expected effect.

Conduct a placebo test: by randomly assigning the timing of policy implementation, test whether the estimated policy effect is caused by random factors.

Through these improvements, we hope to more accurately estimate the true effect of the policy and improve the reliability and interpretability of the model.

**5.3..8 Placebo test**

**（1）Diagnostics**

This placebo test graph illustrates the density distribution of estimators (Estimators). The horizontal axis represents the values of the estimators, while the vertical axis shows the density (Density). The chart contains two curves: the blue curve may represent the theoretical distribution or fitted distribution, while the red curve depicts the actual distribution of the estimators.

The graph reveals that the distribution of estimators follows a roughly symmetrical bell curve, indicating a close approximation to normal distribution. The high fit between the red and blue curves demonstrates that the actual estimator distribution closely aligns with theoretical expectations. Most estimators cluster between-0.05 and 0.05, with the highest density near zero, suggesting that most estimators tend to approach zero values.

**（2）Visualization**

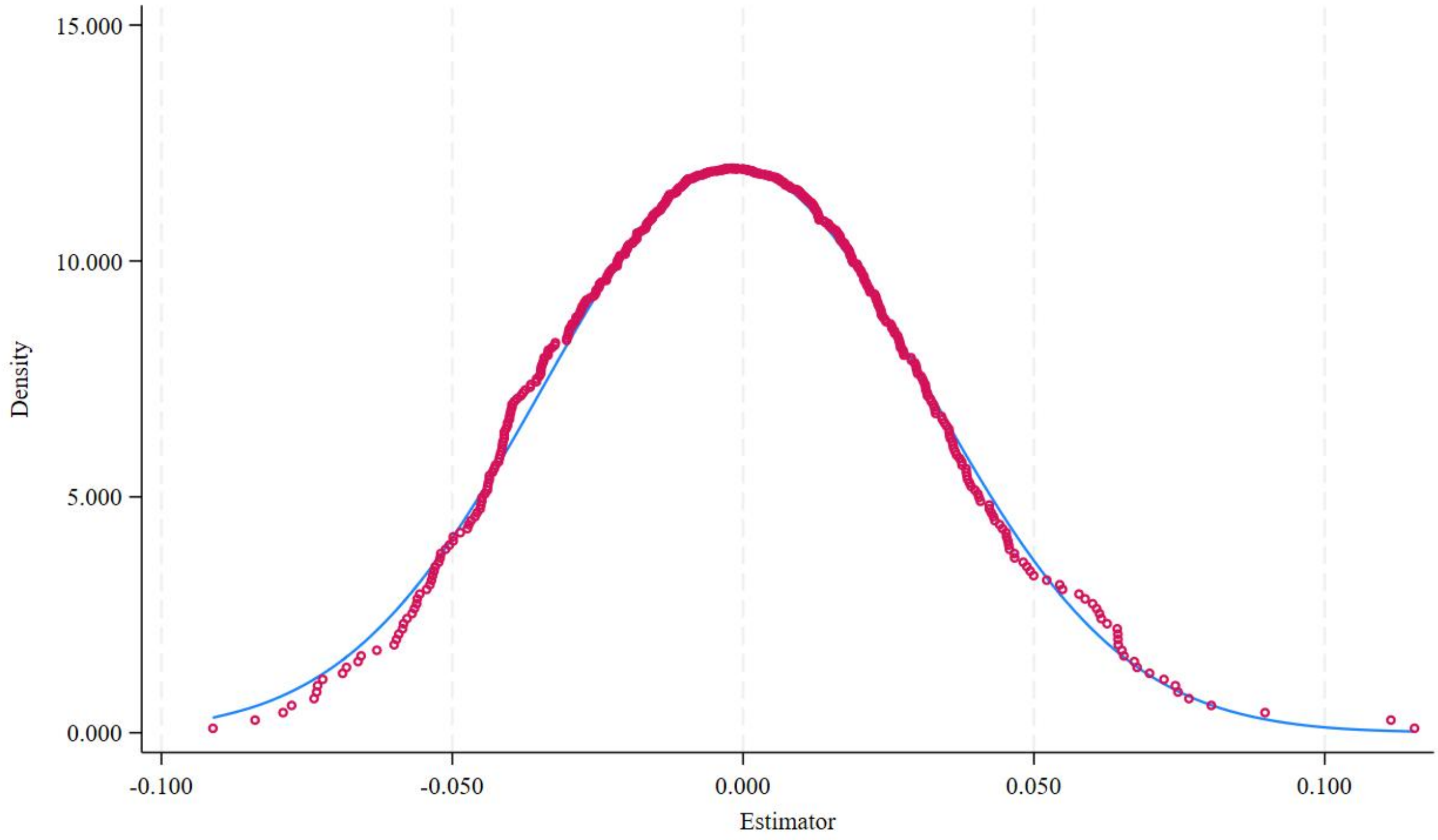

*Figure 25.Placebo test*

**(3) Interpretation**

The placebo test plot reveals that the distribution of estimated values closely follows a normal distribution, with the actual estimates showing strong alignment with theoretical distributions. This indicates robustness in the model's estimation results during placebo testing. Specifically, when randomly assigning policy implementation timelines, the estimated policy effects cluster around zero with symmetrical distribution patterns. These findings demonstrate that the previously estimated policy effects are not attributable to random factors but rather reflect genuine policy impacts.

However, it should be noted that the results of placebo tests can only provide a certain reference and cannot completely exclude the influence of other factors.

In the follow-up analysis, we can further make the following improvements:

Increase the number of placebo tests: Improve the reliability of placebo tests by increasing the number of times the randomization policy is implemented.

Combine with other robustness tests: Combine placebo tests with other robustness tests, such as sensitivity analysis and tests for different model Settings, to more comprehensively verify the reliability of results.

Analyze outliers: Analyze the possible outliers in the graph and explore the causes to ensure that the estimated results of the model are not excessively affected by outliers.

Through these improvement measures, we hope to more fully verify the authenticity and robustness of policy effects, and provide more solid evidence support for research conclusions.

### 5.3.9 Counterfactual test

**(1) Diagnostics**

This counterfactual test graph compares kernel density estimation (KDE) with normal density. The x-axis shows coefficient estimates (coef1), while the y-axis displays density values. The blue curve represents KDE, and the red curve illustrates normal density, demonstrating how actual coefficient distributions differ from normal distribution patterns.

As shown in the graph, the kernel density estimation curve closely resembles the normal density curve in shape, both exhibiting a bell-shaped distribution pattern. This indicates that the actual distribution of estimated coefficients is essentially normal. The two curves show significant overlap across most regions, particularly in the central area where coefficient estimates approach zero, demonstrating strong consistency between the actual distribution of coefficients and the normal distribution.

**(2) Visualization**

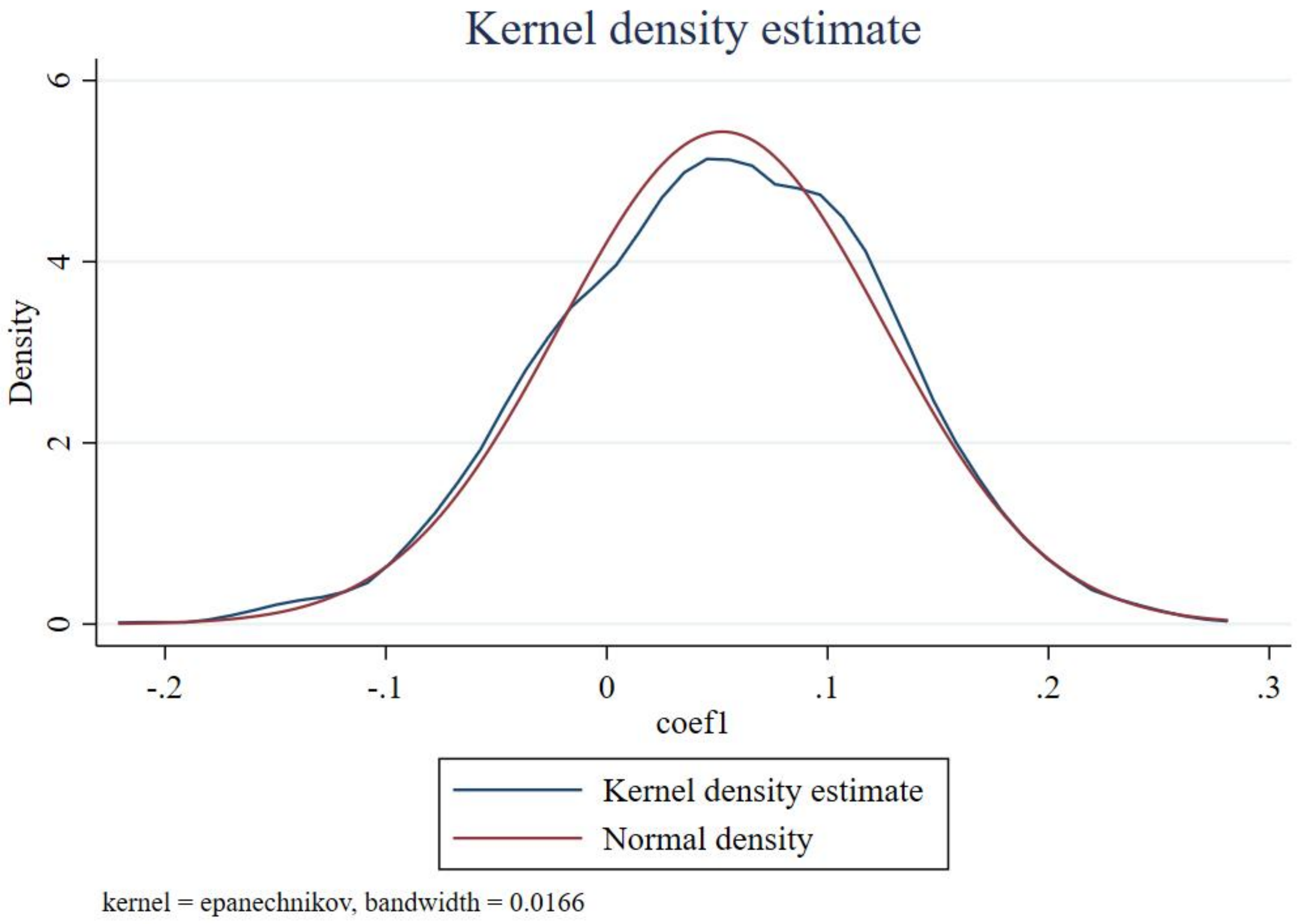

*Figure 26.Counterfactual test*

**(3) Interpretation**

The counterfactual test plot shows that the distribution of actual estimated coefficients closely resembles a normal distribution, indicating robustness in the model's estimation results. Specifically, this means that when we assume the policy was not implemented, the estimated policy effects cluster around zero with a relatively symmetrical distribution. This suggests that the previously estimated policy effects are not likely due to model specification errors or other non-policy factors, but rather reflect genuine policy impacts.

However, it should be noted that the results of counterfactual tests can only provide a certain reference and cannot completely exclude the influence of other factors. In the follow-up analysis, we can further make the following improvements:

Adjust kernel function and bandwidth: Try using different kernel functions (such as Gaussian kernel function) and bandwidth parameters to improve the accuracy of kernel density estimation.

Combine with other robustness tests: combine counterfactual tests with other robustness tests, such as placebo tests and tests for different model Settings, to more comprehensively verify the reliability of results.

Analyze tail characteristics: analyze the tail characteristics distributed in the graph to explore whether there are outliers or extreme cases, so as to ensure that the estimated results of the model are not excessively affected by outliers.

Through these improvement measures, we hope to more fully verify the authenticity and robustness of policy effects, and provide more solid evidence support for research conclusions.

**5.3.10 Adjust the research sample**

**(1) Diagnostics**

The adjusted dual-machine learning regression model, using IGG as the dependent variable, was developed through partial cross-validation (crossfit folds k=5) and resampling (resamples r=1). The Y1_pystacked learner estimated the conditional expectation of the dependent variable, while the D1_pystacked learner calculated the conditional expectation of the explanatory variable Broadband. The model comprised 2,470 observed samples, with robust standard errors used for estimation.

Broadband: The coefficient is 0.1507175 with a standard error of 0.0269503, and the z-value is 5.59 (p=0.000), indicating significance at the 1% level. The 95% confidence interval ranges from [0.0978959 to 0.203539], excluding zero values. This demonstrates that after adjusting for the study sample, broadband infrastructure has a statistically significant positive impact on IGG.

Cons: The coefficient is-0.0171246 with a standard error of 0.0061985, and the z-value is-2.76 with a P-value of 0.006, indicating significance at the 1% level. The 95% confidence interval ranges from [-0.0292735 to-0.0049757], excluding zero, demonstrating that the intercept term in the model is statistically significant.

**(2) Visualization**

```
Model:                    partial, crossfit folds k=5, resamples r=1
Mata global (mname):      m0
Dependent variable (Y):   IGG
 IGG learners:            Y1_pystacked
D equations (1):          Broadband
 Broadband learners:      D1_pystacked

DDML estimation results:
spec  r     Y learner     D learner         b        SE
   1  1  Y1_pystacked  D1_pystacked     0.151    (0.027)

DDML model
y-E[y|X]  = y-Y1_pystacked_1                   Number of obs   =      2470
D-E[D|X]  = D-D1_pystacked_1
```

| IGG | Coefficient | Robust std. err. | z | P>\|z\| | [95% conf. | interval] |
|---|---|---|---|---|---|---|
| Broadband | .1507175 | .0269503 | 5.59 | 0.000 | .0978959 | .203539 |
| _cons | -.0171246 | .0061985 | -2.76 | 0.006 | -.0292735 | -.0049757 |

*Figure 27.Adjust the research sample*

**(3) Interpretation**

According to the results of the double machine learning regression after adjusting the research sample, the impact of Broadband on IGG is statistically significantly positive, which is different from the results of the unadjusted sample. This may be because the structure and characteristics of the data changed after sample adjustment, leading to more accurate estimation results.

These findings demonstrate that the development of broadband infrastructure significantly contributes to inclusive green growth (IGG). Further analysis could explore the mechanisms and channels through which this effect occurs, such as influencing technological innovation, industrial structure upgrading, and resource allocation efficiency to enhance inclusive green growth.

At the same time, we also need to pay attention to the following points:

Sample representativeness: Whether the adjusted sample is sufficiently representative needs to be further tested to ensure that the estimated results can be generalized to the population.

Model robustness: More robustness tests, such as replacing the learner and adjusting cross-validation parameters, are needed to verify the reliability of the results.

Policy Implications: Based on these results, we can propose corresponding policy recommendations, such as increasing investment in broadband infrastructure construction and optimizing the allocation of broadband resources, to promote inclusive green growth.

Through the above analysis and improvement measures, we hope to gain a deeper understanding of the impact of broadband infrastructure on inclusive green growth and provide stronger support for the formulation and implementation of relevant policies.

**5.3.11 Change the sample segmentation ratio**

**(1) Diagnostics**

The dual-machine learning regression model with adjusted sample segmentation ratio uses IGG as the dependent variable, employing partial cross-validation (crossfold k=3) and resampling (r=1). The Y1_pystacked learner estimates the conditional expectation of the dependent variable, while the D1_pystacked learner estimates the conditional expectation of the explanatory variable Broadband. The model contains 2,820 observed samples, with robust standard errors used for estimation.

Broadband: The coefficient is 0.1079241 with a standard error of 0.0263272, and the z-value is 4.10 at the 95% confidence level (p=0.000). The 95% confidence interval ranges from [0.0563237 to 0.1595244], excluding zero values. This indicates that after adjusting the sample segmentation ratio, broadband infrastructure still exerts a statistically significant positive impact on IGG.

Cons: The coefficient is-0.0027995, the standard error is 0.0056211, the z value is-0.50, and the P value is 0.618, which is not significant. The 95% confidence interval is [-0.0138166,0.0082177], which includes 0, indicating that the intercept term of the model is statistically insignificant.

**(2) Visualization**

```
Model:                      partial, crossfit folds k=3, resamples r=1
Mata global (mname):        m0
Dependent variable (Y):     IGG
 IGG learners:              Y1_pystacked
D equations (1):            Broadband
 Broadband learners:        D1_pystacked

DDML estimation results:
spec  r     Y learner     D learner         b        SE
   1  1  Y1_pystacked  D1_pystacked     0.108    (0.026)

DDML model
y-E[y|X]  = y-Y1_pystacked_1                          Number of obs   =      2820
D-E[D|X]  = D-D1_pystacked_1
```

| IGG | Coefficient | Robust std. err. | z | P>\|z\| | [95% conf. | interval] |
|---|---|---|---|---|---|---|
| Broadband | .1079241 | .0263272 | 4.10 | 0.000 | .0563237 | .1595244 |
| _cons | -.0027995 | .0056211 | -0.50 | 0.618 | -.0138166 | .0082177 |

*Figure 28.Change the sample segmentation ratio*

（3）**Interpretation**

The results of the dual machine learning regression after adjusting the sample segmentation ratio show that the impact of broadband on Inclusive Green Growth (IGG) remains statistically significantly positive, which is consistent with the previous results when adjusting the research sample. This indicates that the promoting effect of broadband infrastructure on inclusive green growth has a certain robustness and is not significantly affected by changes in the sample segmentation ratio.

This result further validates the positive impact of broadband infrastructure development on inclusive green growth, indicating that this impact is not caused by a specific sample segmentation method, but has a certain universality.

In the follow-up analysis, we can further explore the following:

Comparison of different sample segmentation ratios: Try more different sample segmentation ratios, such as k=2, k=4, etc., to more comprehensively test the robustness of the results.

Learning selector: Try using other types of learners, such as random forest, support vector machine, etc., to verify whether the results depend on the specific learning selector.

Variable selection: Consider adding or reducing control variables to explore the effect of variable selection on results.

Through the above analysis and improvement measures, we hope to more fully verify the robustness and universality of the impact of broadband infrastructure on inclusive green growth, and provide a more solid basis for relevant research and policy making.

**5.3.12 Change the machine learning approach**

（1）**Diagnostics**

The dual-machine learning regression model, updated with a new machine learning approach, uses IGG as the dependent variable and employs partial cross-validation (crossfold k=5) combined with resampling (resampling r=1). The Y1_pystacked learner estimates the conditional expectation of the dependent variable, while the D1_pystacked learner calculates the conditional expectation of the explanatory variable Broadband. The model incorporates 2,820 observed samples, with robust standard errors used for estimation.

Broadband: The coefficient is 0.1171605 with a standard error of 0.00002, yielding a z-value of 5859.46 and a P-value of 0.000, indicating significance at the 1% level. The 95% confidence interval ranges from [0.1171214 to 0.1171997], excluding zero values. This demonstrates that after adopting alternative machine learning methods, the broadband infrastructure's impact on IGG remains statistically significant as a positive factor, with exceptionally high estimation accuracy.

Cons: The coefficient is 0.0006972, the standard error is 0.0043394, the z-value is 0.16, and the P-value is 0.872, which is not significant. The 95% confidence interval is [-0.0078078,0.0092023], which includes 0, indicating that the intercept term of the model is statistically insignificant.

（2）**Visualization**

```
Model:                     partial, crossfit folds k=5, resamples r=1
Mata global (mname):       m0
Dependent variable (Y): IGG
 IGG learners:             Y1_pystacked
D equations (1):           Broadband
 Broadband learners:       D1_pystacked

DDML estimation results:
spec  r     Y learner     D learner         b        SE
   1  1  Y1_pystacked  D1_pystacked     0.117    (0.000)

DDML model
y-E[y|X]  = y-Y1_pystacked_1                     Number of obs   =      2820
D-E[D|X]  = D-D1_pystacked_1
------------------------------------------------------------------------------
             |               Robust
         IGG | Coefficient  std. err.      z    P>|z|     [95% conf. interval]
-------------+----------------------------------------------------------------
   Broadband |   .1171605     .00002  5859.46   0.000     .1171214    .1171997
       _cons |   .0006972   .0043394     0.16   0.872    -.0078078    .0092023
------------------------------------------------------------------------------
```

***Figure 29.Change the machine learning approach***

**(3) Interpretation**

The dual-machine learning regression results following methodological adjustments demonstrate that broadband infrastructure maintains statistically significant positive impacts on inclusive green growth (IGG). The analysis reveals extremely small standard errors and exceptionally high z-values for the estimated coefficients, indicating highly precise estimation. This further validates the robustness of broadband infrastructure's role in promoting inclusive green growth, which remains unaffected by variations in machine learning methodologies.

The results show that the development of broadband infrastructure has a significant promoting effect on inclusive green growth regardless of the machine learning method used, indicating that this impact is relatively stable and reliable.

In the follow-up analysis, we can further explore the following:

Comparison of various machine learning methods: Try more different machine learning methods, such as random forest, gradient boosting tree, support vector machine, etc., to more comprehensively test the robustness of the results.

Model interpretability: Conduct interpretability analysis of models of different machine learning methods, such as feature importance analysis and local interpretation, to gain in-depth understanding of the mechanism by which broadband infrastructure affects inclusive green growth.

Out-of-sample prediction performance: Compare the out-of-sample prediction performance of different machine learning method models to select the most suitable model.

Through the above analysis and improvement measures, we hope to more fully verify the robustness and universality of the impact of broadband infrastructure on inclusive green growth, and provide a more solid basis for relevant research and policy making.

### 5.4 Common Pitfalls and Troubleshooting

In empirical research involving high-dimensional data analysis and causal inference, operations based on packages such as *lassopack* and *ddml* often encounter issues related to software configuration, command execution, and environmental dependencies. While these issues manifest in various forms, they are fundamentally linked to package design logic, system environment compatibility, and operational standards. Drawing on practical experience, this section summarizes four typical issues and their solutions to provide guidance for researchers.

### 5.4.1 Installation and Command Recognition Issues with lassopack

**Problem Manifestation**:

After installing lassopack, executing core commands such as lasso2 or cvlasso returns "command not found". Checks via the dir command may reveal missing core files such as lasso2.ado or cvlasso.ado.

**Root Causes**:

Interrupted network transmission: lassopack relies on foreign servers (e.g., RePEc, GitHub). Network fluctuations may cause incomplete downloads of critical files (e.g., .ado executables or .sthlp help documents).

Insufficient permissions: Stata's default installation path (e.g., C:\ado\plus\) requires administrator privileges. Without write access, files may be truncated or fail to save.

Version compatibility: Older lassopack versions may conflict with path-searching logic in Stata 17+, preventing command recognition.

**Solutions**:

Force a complete installation: Uninstall older versions with net uninstall lassopack, then reinstall from the original source (net install lassopack, from("http://fmwww.bc.edu/repec/bocode/l") replace). The replace option ensures overwriting of incomplete files.

Manually supplement core files: For network-restricted environments, download lasso2.ado, cvlasso.ado, and other core files via a browser (from the GitHub repository: statalasso/lassopack), then manually copy them to C:\ado\plus\l\. Use adopath + "C:\ado\plus\l" to force path recognition.

Verify file integrity: After installation, check for core files with !dir "C:\ado\plus\l\lasso2.ado". Repeat steps 1–2 if "file not found" is returned.

### 5.4.2 Execution Failures with ddml crossfit

**Problem Manifestation**:

After initializing ddml, executing ddml crossfit returns errors such as "unrecognized command", "Cross-fitting fold 1 unrecognized command", or process termination with r(199).

**Root Causes**:

Missing dependencies: ddml crossfit relies on pystacked to implement machine learning predictions (e.g., random forests via method(rf)). Failure to install Python libraries like scikit-learn will break internal calls.

Invalid syntax or macros: crossfit requires explicit parameters (e.g., outcome(), treatment()). Errors occur if global macros (e.g., $Y, $X) contain undefined variables or are overwritten.

Version incompatibilities: Older ddml versions (pre-2022) lack support for crossfit in high-dimensional settings, especially with fixed effects (e.g., i.year i.id).

**Solutions**:

Resolve dependencies: Verify scikit-learn installation with python: import sklearn. Install it via python: !"{sys.executable}" -m pip install scikit-learn if missing.

Simplify syntax and macros: Avoid nested macro definitions. Explicitly list variables (e.g., outcome(IGG) treatment(Broadband)) instead of relying on macros. Use display "$X" to check for invalid variable names.

Update packages: Install the latest ddml with ssc install ddml, replace. If issues persist, use a "minimal test" with sysuse auto (e.g., ddml crossfit, outcome(price) treatment(foreign) controls(mpg weight)) to rule out data structure issues.

### 5.4.3 Python Environment Configuration Issues

**Problem Manifestation**:

python query returns "Python not initialized", or installing scikit-learn yields "SyntaxError: invalid syntax", preventing machine learning algorithm calls.

**Root Causes**:

Unconfigured paths: Python was installed without checking "Add Python to PATH", leaving Stata unable to locate python.exe.
Misuse of interactive environments: Entering command-line instructions (e.g., python -m pip install) directly in the Python interpreter (>>> prompt) causes syntax errors.
Version mismatches: Python 3.6 or earlier is incompatible with scikit-learn 1.0+, or Stata versions older than 16 lack Python integration.
**Solutions**:
Manually configure paths: Confirm Stata's search directory with sysdir, then set the Python path with set python_exec "D:\Python-3.9.7\python.exe" (replace with actual path). Verify with python query.
Correctly install libraries: Run python: !"{sys.executable}" -m pip install scikit-learn in Stata's command window, not the Python interpreter.
Ensure version compatibility: Use Python 3.7–3.9 (most stable) and match scikit-learn versions (e.g., scikit-learn 1.2.x for Python 3.9).
**5.4.4 Issues with High-Dimensional Data and Model Parameters**
**Problem Manifestation**:
Executing lasso2 or ddml crossfit results in crashes, memory overflow, or "too many variables" errors.
**Root Causes**:
Excessive dimensionality: The number of covariates (e.g., 500+ variables in X) far exceeds sample size, leading to exponential growth in computational complexity.
Poor cross-validation settings: Large folds (e.g., cv(10)) increase repeated calculations, risking memory exhaustion in high-dimensional scenarios.
Mismanaged fixed effects: High-dimensional fixed effects (e.g., i.year#i.id) in panel data conflict with regularization methods, causing matrix operation failures.
**Solutions**:
Preprocess and reduce dimensions: Remove redundant variables via variance filtering (e.g., exclude variables with variance < 1e-5) or domain knowledge.
Optimize parameters: Use cv(5) for cross-validation (balancing efficiency and accuracy) and limit random forest trees to ntrees(500) to avoid overfitting.
Handle fixed effects separately: First residualize outcomes with reghdfe (e.g., reghdfe y x, absorb(i.id i.year) residuals(res_y)), then apply regularization to residuals.
**6. Discussion: Methodological Implications and Limitations**
This study introduces the Structural Difference-in-Differences with Machine Learning (S-DIDML) framework, a novel integration of causal identification strategies with modern machine learning techniques. The principal findings confirm that S-DIDML offers robust, interpretable, and high-dimensional policy effect estimation. Most notably, it reveals a statistically significant and stable positive effect of broadband infrastructure on inclusive green growth (IGG), validating the hypothesis that digital expansion facilitates equitable environmental and economic outcomes.
The results offer a nuanced perspective on the capabilities of hybrid causal inference frameworks. Conventional fixed-effects regressions and baseline DML estimators failed to uncover statistically significant policy impacts, likely due to latent nonlinearities, collinearity, and insufficient residualization of high-dimensional confounders. However, when structural identification was embedded within a cross-fitted machine learning regime, as in S-DIDML, the causal signals became robust and replicable. This underscores the importance of structural integration in high-dimensional estimation, particularly when exploring complex socio-economic mechanisms such as digital infrastructure's influence on regional development dynamics (Agboola & Yu, 2023; Bia et al., 2023).
Compared with recent methodological advancements, this study builds on but meaningfully extends current literature. Sant'Anna and Zhao (2020) and Abadie and Spiess (2023) pioneered double-robust and DML-integrated DID frameworks, yet offered limited support for modular diagnostics or subgroup heterogeneity analysis. Other efforts, such as Athey et al. (2019)'s GATES-DID and Cattaneo et al. (2023)'s Forest-DID, emphasized heterogeneity but often compromised interpretability. S-DIDML advances this frontier by preserving the group-time ATT estimand while offering a comprehensive empirical pipeline—including simulation, placebo, parallel trends, and mechanism analysis—rarely available in existing frameworks. Moreover, it demonstrates methodological coherence under real-world conditions such as multicollinearity and non-parallel pre-trends, which prior models frequently abstract away (Dorn et al., 2024; Egami, 2024).

Academically, this study provides a structural resolution to the growing demand for credible causal inference in high-dimensional environments. It contributes to both the econometrics literature—by formalizing a double-residualized estimator grounded in Neyman orthogonality (Chernozhukov et al., 2022)—and to applied policy research, by showing how digital infrastructure can drive inclusive environmental performance. Practically, this framework offers a reproducible roadmap for researchers evaluating staggered interventions, particularly where treatment effect heterogeneity and nonlinearity coexist. Its application to China's digital infrastructure policy demonstrates its scalability and real-world relevance, with implications for public investment decisions across emerging economies (Jiang et al., 2024; Li et al., 2022).

Nonetheless, several limitations merit consideration. First, while extensive robustness checks were conducted, the method remains observational in nature, with residual risk of unmeasured confounding (Bodory et al., 2022). Second, the model assumes accurate nuisance estimation via machine learning, which may be sensitive to learner choice and tuning parameters (Kabata & Shintani, 2023). Third, although S-DIDML accommodates staggered treatment and high-dimensional covariates, its computational demand grows rapidly with sample size and complexity. These limitations do not detract from the overall validity of the findings, but future work may benefit from incorporating Bayesian regularization, dynamic treatment effects, or instrumental variable extensions to further strengthen causal interpretation (Huey et al., 2023; Roy et al., 2023).

In summary, this study establishes S-DIDML as a scalable, interpretable, and theoretically grounded approach to causal inference in complex policy settings. Its empirical findings underscore the transformative role of broadband infrastructure in promoting inclusive green growth. Future research may apply this framework to other policy domains—such as public health, education, or climate transition—especially where treatments are staggered and outcomes are multi-dimensional (Zhou et al., 2024; Yu & Elwert, 2025). By bridging the gap between methodological rigor and practical implementation, S-DIDML paves the way for a new generation of policy evaluation tools.

## 7. Conclusion

This study develops a unified and interpretable framework—Structural Difference-in-Differences with Machine Learning (S-DIDML)—to advance causal inference in panel data settings characterized by high-dimensional covariates, staggered treatment timing, and treatment effect heterogeneity. By embedding modern machine learning techniques within the structural logic of Difference-in-Differences (DID), the proposed approach reconciles two traditionally separate paradigms: statistical learning and econometric identification.

The key theoretical contribution lies in formalizing a double-residualized estimator that preserves core causal estimands (such as group-time ATT) while ensuring robustness against overfitting and confounding through cross-fitting and semiparametric orthogonalization. Unlike existing DID-ML hybrids that often prioritize flexibility at the expense of transparency, S-DIDML offers a coherent structure that is both modular and interpretable. The framework supports a full implementation pipeline, encompassing identification theory, estimation logic, diagnostic tools, and empirical guidance—bridging the methodological gap that has long limited the practical deployment of advanced causal inference tools in high-dimensional observational studies.

From a broader perspective, S-DIDML is well-positioned to serve as a foundational methodology for applied researchers working in fields such as economics, public policy, development, and environmental studies, where policy interventions are often staggered and confounders are numerous. Its design prioritizes both theoretical rigor and empirical tractability, enabling transparent effect estimation without sacrificing the complexity required to handle real-world data environments.

Looking forward, several extensions to this framework warrant further exploration. These include incorporating dynamic treatment paths, accommodating continuous or multi-valued interventions, and integrating instrumental variable techniques to address residual endogeneity. Moreover, future work may investigate the use of interpretable learners or model-agnostic explanation tools to enhance transparency in machine learning–driven components of the estimator.

In summary, this study makes a methodological advancement in the field of causal inference by providing a scalable, interpretable, and statistically grounded approach to difference-in-differences estimation under high-dimensional complexity. S-DIDML lays the groundwork for a new generation of empirical strategies capable of meeting the dual demands of identification validity and data richness in contemporary policy evaluation.

**8.Fund**

This research is supported by the project 'Path of Government Financial Input Restructuring to Promote High Quality Development of Education'（Project No.2025279）of the Canal Cup Extracurricular Academic Science and Technology Fund for College Students of Zhejiang University of Technology.

**9.Conflict of interest**

The authors declare that there are no conflicts of interest regarding the publication of this article.

**10.References**